\documentclass[10pt]{article}
\usepackage{style}

\title{Optimal Transport Dropout for Structured Predictive Uncertainty}

\author[1]{Giacomo Lorenzon\thanks{giacomo.lorenzon@polimi.it}}
\author[1]{Francesco Regazzoni}
\affil[1]{MOX--Department of Mathematics, Politecnico di Milano, Italy}

\date{}

\begin{document}
\maketitle

\begin{abstract}
Deterministic neural networks and neural operators provide point predictions with no intrinsic measure of reliability. Yet, predictive uncertainty may stem from irreducible outcome variability, finite data, or limitations of the chosen model class. Monte Carlo dropout offers a computationally convenient way to construct a predictive distribution through stochastic feature masking, without training multiple independent networks or explicitly inferring a posterior over model parameters. However, its perturbation law is largely prescribed \textit{a priori} and typically factorised across latent coordinates. We introduce \textit{Optimal Transport Dropout} (OTD), which instead learns the predictive mapping and the law of its latent perturbations jointly. Starting from a simple independent reference distribution, OTD transports latent perturbations through a learnable flow and propagates them through the predictive neural network, thereby inducing a structured predictive law. Training uses the strictly proper Energy Score, while a kinetic-action term geometrically regularises the transport. Synthetic benchmarks show that OTD captures multimodal predictive distributions, generates meaningful dispersion when the model is misspecified, and exhibits contracting dispersion as more training data or greater model capacity are provided. For a field-valued partial differential equation surrogate, predictive dispersion strongly aligns with the spatial pattern of prediction errors. On this task, compared with Monte Carlo dropout, OTD yields more accurate predictions and better-calibrated, substantially narrower intervals. On real-world regression benchmarks, it further shows competitive accuracy and better probabilistic predictions compared to several established baselines. OTD therefore offers a way to learn structured predictive uncertainty without explicit posterior inference or ensembles of independently trained predictors.
\end{abstract}

\section{Introduction}

Machine-learning models can provide accurate predictions across a wide range of applications, but standard deterministic architectures provide no intrinsic measure of predictive reliability \cite{he2025survey, quarteroni2025combining}. Uncertainty may arise from intrinsic variability or from limitations of the learned model. Within the latter, we distinguish uncertainty due to finite observations from that associated with model-class misspecification \cite{he2025survey, murphy2023probabilistic}. We refer to these sources as aleatoric, finite-data epistemic, and model-class uncertainty, respectively. Aleatoric uncertainty is irreducible and reflects variability in the conditional target distribution; finite-data epistemic uncertainty arises because the available observations only partially constrain the input--output relation; model-class uncertainty arises when the chosen model family cannot represent the target relation exactly. In practice, these sources may coexist \cite{he2025survey} and are not always separately identifiable. In this work, we focus on learning their combined effect through a predictive distribution rather than explicitly decomposing them.

Scientific surrogate modelling is a particularly relevant setting for this problem. In this context, neural surrogates are machine-learning models trained to approximate the input-output map of computationally expensive high-fidelity simulations, providing substantially cheaper evaluations \cite{quarteroni2025combining}. Since the surrogate is deployed in place of the underlying solver, inaccurate predictions may occur without an explicit indication of failure \cite{murphy2023probabilistic}. Reliable uncertainty estimates are therefore needed to identify predictions that should not be trusted, while preserving the computational advantage that motivates the use of the surrogate itself.

Bayesian learning provides a principled framework for epistemic uncertainty by placing a distribution over model parameters and inducing a posterior predictive distribution \cite{murphy2023probabilistic,kendall2017uncertainties,pmlr-v37-blundell15, derkiureghian2009aleatory}. Full Bayesian inference, however, is typically computationally demanding \cite{tomczak_efcient, pmlr-v48-louizos16, mackay1992practical}. Monte Carlo (MC) dropout provides a lightweight alternative based on a single network and stochastic forward passes, while admitting an approximate Bayesian interpretation \cite{gal2017concrete, pmlr-v48-gal16}. Despite its efficiency, this approach retains a perturbation law whose structure is largely prescribed a priori, typically through factorised masks.

This restriction motivates a connection with generative modelling, where flexible probability laws are learned from simple reference distributions \cite{goodfellow2016deep}. A closely related limitation arises in variational inference, where restrictive approximate posterior families may fail to represent strongly non-Gaussian or correlated distributions \cite{tomczak_efcient, pmlr-v48-louizos16, pmlr-v37-rezende15}. Normalising flows address this limitation by transporting a simple reference law through learnable transformations \cite{ziegler_latent, pmlr-v70-louizos17a, kingma2016improved, pmlr-v37-rezende15}. This suggests a complementary approach to uncertainty quantification: rather than prescribing the stochastic perturbation law, one may learn how a simple reference distribution should be transformed before being propagated through the predictive network.

We introduce \textit{Optimal Transport Dropout} (OTD), in which a latent masking variable is transported through a learnable flow, and subsequently propagated through the predictive network. The transported law can encode dependence, anisotropy, and multimodality across mask coordinates, extending the perturbation family beyond factorised masks. The induced predictive distribution is trained directly using the Energy Score \cite{szekely2023energy}, while a kinetic action geometrically regularises the transport \cite{peyre2019computational, ambrosio_users_2013, villani2009optimal}. OTD therefore retains the lightweight stochastic structure of MC dropout while replacing its prescribed perturbation law with a learnable structured distribution. OTD targets structured predictive uncertainty rather than a single source of uncertainty. For deterministic targets, its predictive dispersion is investigated as a signal of finite-data and model-class uncertainty; for stochastic targets, the predictive law additionally represents aleatoric variability. When these effects coexist, OTD learns their combined predictive distribution without requiring an explicit decomposition. We call the resulting uncertainty \emph{structured} because the transported perturbation law can encode dependencies across mask coordinates that are absent from standard factorised dropout. We assess the properties of OTD through controlled experiments isolating distributional expressiveness, model misspecification, model capacity, and data availability, before considering field-valued scientific surrogates and real-world regression problems.

\section{Related work}

\paragraph{Bayesian and stochastic uncertainty quantification.}
Bayesian neural networks model parameter uncertainty through approximate posterior inference \cite{murphy2023probabilistic, pmlr-v37-blundell15}. MC dropout interprets stochastic masking as approximate Bayesian inference \cite{pmlr-v48-gal16}, while Concrete Dropout learns dropout probabilities \cite{gal2017concrete}. GFlowOut goes beyond independent masks by learning a potentially multimodal and correlated posterior distribution over dropout masks using generative flow networks \cite{pmlr-v202-liu23r}. Deep ensembles use independently trained predictors \cite{he2025survey}, whereas Epinets augment a predictor with an auxiliary stochastic network \cite{osband2023epistemic}. Stochastic flow matching has also been combined with approximate Bayesian inference and antithetic sampling for uncertainty estimation in scientific imaging \cite{wu2026uncertainty}. OTD addresses a different optimisation problem: it transports a reference law into a continuous latent perturbation law and trains the induced predictive distribution directly with the Energy Score, subject to kinetic-action regularisation. Its objective therefore does not require a likelihood, a variational posterior over masks, or a trajectory-balance loss.

\paragraph{Generative modelling and transport.}
Normalising flows enrich simple reference laws through learnable invertible maps \cite{pmlr-v37-rezende15}; related work develops structured and flow-based posterior approximations \cite{pmlr-v70-louizos17a, pmlr-v48-louizos16, kingma2016improved}. Conditional generative models similarly learn predictive push-forward distributions \cite{franco_nonparametric_2025}. OTD applies this principle in perturbation space, transporting a relaxed dropout law through a kinetically regularised flow. This differs from post-hoc aggregation of uncertainty scores through Monge--Kantorovich ranks \cite{kotelevskii2025multidimensional}, which operates on scores rather than the law inducing predictions.

\paragraph{Uncertainty in functional surrogates.}
Probabilistic Fourier neural operators learn distributions over solution fields using scoring rules \cite{bulte2024probabilistic}, while structure-aware methods examine how perturbation placement and form affect uncertainty estimates \cite{song_structure_aware_2026}. REEF-GP fits a Gaussian process to residuals of a frozen neural operator using its internal representations \cite{vendrellgallart2026geometry}. GenUQ generates a subset of operator parameters with a hyper-network trained using the Energy Score \cite{yen2025genuq}. OTD instead learns correlated feature perturbations, rather than fitting a post-hoc residual model or generating operator parameters, and is not tied to a particular neural-operator architecture.

\paragraph{Scoring-rule training.}
Proper scoring rules provide principled objectives for learning predictive distributions \cite{gneiting2007strictly}. The Energy Score admits sample-based estimation without requiring a tractable predictive density \cite{szekely2023energy}, thus enabling training of implicit generative models \cite{pacchiardi2024score}. OTD uses this formulation to train its induced output law and can be viewed as structured engression, with noise entering through transported masks. Engression error analyses distinguish approximation, statistical, and Monte Carlo contributions \cite{chen2026error}; OTD additionally involves transport discretisation and kinetic regularisation. Work on scale-sensitive scoring motivates complementary spatial diagnostics for field-valued predictions \cite{lang2026sensitivity}.

\paragraph{Interpretation of predictive uncertainty.}
Predictive dispersion need not capture model bias or the joint effects of dataset and procedural variability \cite{jimenez2025epistemic}, and differs from predictive performance and decision-relevant uncertainty \cite{pmlr-v267-bickford-smith25a}. In generative models, sample variability may additionally mix intrinsic and epistemic components, motivating methods that estimate epistemic uncertainty separately from generative dispersion \cite{pmlr-v300-gupta26b}. Risk-based formulations further combine posterior variability with predictive bias \cite{wizgall2026unified}. Accordingly, we interpret OTD dispersion as a learned predictive quantity whose epistemic content must be established through reducibility, error-alignment, and misspecification diagnostics, rather than identifying it a priori with epistemic uncertainty.

\section{Methods}

\subsection{Optimal Transport Dropout: from independent to transported masks}
\paragraph{Predictor and stochastic modulation.} Let $\mathcal D=\{(x_i,y_i)\}_{i=1}^n$ be observations from an unknown conditional law $Y\mid X=x\sim P_x^\ast$. Consider now a neural network acting from $\mathcal X$ to $\mathcal Y$. For notational simplicity, consider a single hidden layer network with hidden representation $h_\theta(x)\in\mathbb R^{d_z}$. Given a random mask $Z\in[0,1]^{d_z}$, stochastic modulation replaces $h_\theta(x)$ with $Z\odot h_\theta(x)$. We denote the resulting prediction by $\hat Y=G_\theta(x,Z)$. The extension to multiple modulated layers is described in \autoref{sec:appendix_implementation}.

\paragraph{Learnable mask transport.}
Monte Carlo dropout samples independent Bernoulli mask coordinates \cite{pmlr-v48-gal16}. OTD starts from a factorised reference law, $Z_0\sim\rho_0$, $\rho_0=\otimes_{\ell=1}^{d_z}\rho_{0,\ell},$ where each $\rho_{0,\ell}$ is chosen as a continuous relaxation of a Bernoulli law. It then learns a transport map $T_\psi:[0,1]^{d_z}\to[0,1]^{d_z}$, parametrised through a neural flow, such that  $Z_1=T_\psi(Z_0)$. Its implementation is detailed in \autoref{sec:appendix_implementation}. The transported mask is thus applied to the hidden representation, yielding $\hat Y=G_\theta(x,Z_1)$.

\paragraph{Induced predictive distribution.} In other words, the transport transforms the reference law into $\rho_1^\psi=(T_\psi)_\#\rho_0,$ which induces the predictive distribution
$$
P^{\theta,\psi}(\cdot\mid x)
=
\bigl(G_\theta(x,\cdot)\bigr)_\#
\rho_1^\psi.
$$
Although $\rho_1^\psi$ is shared across inputs, propagating it through the input-dependent map $G_\theta(x,\cdot)$ yields a predictive distribution specific to each $x \in \mathcal X$. The transport can introduce dependence across mask coordinates, extending the perturbation family beyond the factorised reference law.

\subsection{Learning objective}
\paragraph{Learning the transport.}
The terminal mask distribution is not specified in advance. Instead, the transport network parameters $\psi$ and the predictive network parameters $\theta$ are learned jointly from data. The predictive law induced by the transported masks is optimised through the Energy Score \cite{szekely2023energy}, while the latent flow is regularised through its kinetic action \cite{ambrosio_users_2013}. Consequently, the data-fitting term constrains the perturbation distribution only through its effect on the resulting predictions, while the kinetic action regularises the transport directly.

\paragraph{Energy Score.} OTD is trained at the level of predictive distribution rather than through a pointwise regression loss. Let $P$ and $Q$ be probability distributions on the output space $\mathcal Y$, namely $P,Q \in \mathcal P_1(\mathcal  Y)$. Their Energy Distance \cite{szekely2023energy} is defined as
\begin{equation}
\mathcal E_d(P,Q) = 2\,\mathbb E_{Y\sim P,\;V\sim Q} d_{\mathcal Y}(Y,V) - \mathbb E_{Y,Y'\overset{\mathrm{iid}}{\sim}P} d_{\mathcal Y}(Y,Y') - \mathbb E_{V,V'\overset{\mathrm{iid}}{\sim}Q} d_{\mathcal Y}(V,V'),
\label{eq:energy_distance}
\end{equation}
where $d_{\mathcal Y}$ is a metric of strong negative type (\cite{szekely2023energy}, Definition 10.4). Under this condition, $\mathcal E_d(P,Q)\geq 0$, with equality if and only if $P=Q$ \cite{bulte2024probabilistic, szekely2023energy}.

In supervised learning, however, the target distribution $Q = P_x^\ast$, for fixed $x \in \mathcal X$, is not directly available: for each input $x$, we observe only a realisation $y\sim Q$. Since the last term in \autoref{eq:energy_distance} does not depend on the predictive distribution $P$, minimising the expected Energy Distance with respect to $P$ is equivalent to minimising the Energy Score
\begin{equation*}
S_{\mathrm{ES}}(P,y)
=
\mathbb E_{Y\sim P}
d_{\mathcal Y}(Y,y)
-
\frac12
\mathbb E_{Y,Y'\overset{\mathrm{iid}}{\sim}P}
d_{\mathcal Y}(Y,Y').
\label{eq:energy_score}
\end{equation*}
More precisely, if $V\sim Q$,
$$
\mathbb E_{V\sim Q}
\left[
S_{\mathrm{ES}}(P,V)
\right]
-
\mathbb E_{V\sim Q}
\left[
S_{\mathrm{ES}}(Q,V)
\right]
=
\frac12\,\mathcal E_d(P,Q).
$$
Hence, the Energy Score is also strictly proper whenever $d_{\mathcal Y}$ is of strong negative type. For the deterministic setting, $Q=\delta_{y(x)}$, and therefore $S_{\mathrm{ES}}(P,y(x))=\frac12\,\mathcal E_d(P,\delta_{y(x)}).$ \autoref{sec:appendix_es} collects some results regarding optimisation under the Energy Score.

\paragraph{Dynamic formulation and kinetic regularisation.}
The transport is generated by a time-dependent velocity field $v_\psi: [0,1] \times \mathcal Z \to \mathbb R^{d_z} $. In the chosen transport coordinates, trajectories satisfy
$$
\frac{\mathrm dz_t}{\mathrm dt}
=
v_\psi(t,z_t),
\qquad t\in[0,1].
$$
Under standard well-posedness assumptions, their marginal laws $\mu_t$ satisfy the continuity equation
$$
\partial_t\mu_t
+
\nabla\cdot(\mu_t v_\psi)
=
0
$$
in the weak sense. The parametrisation relating these trajectories to the bounded masks is detailed in \autoref{sec:appendix_implementation}. Different transport paths may induce similar predictive distributions. We therefore regularise the learned flow through its kinetic action,
$$
\mathcal A(\psi)
=
\frac12
\int_0^1
\mathbb E_{z\sim\mu_t}
\|v_\psi(t,z)\|_2^2
\,\mathrm dt.
$$
This is the action underlying the dynamic formulation of quadratic optimal transport \cite{peyre2019computational, ambrosio_users_2013, villani2009optimal}. For prescribed endpoint laws with finite second moments, its infimum over all admissible paths equals half their squared Wasserstein distance. Consequently, any admissible parametrised path satisfies
$$
\frac12 W_2^2(\mu_0,\mu_1)
\leq
\mathcal A(\psi).
$$
OTD does not explicitly solve an optimal transport problem between prescribed endpoints, but it penalises the action of its learned path while selecting the terminal mask law through the predictive objective. The specific role of the penalisation term is investigated in \autoref{sec:appendix_kinetic}.

\paragraph{Training objective.} Given training observations $\mathcal D = \{(x_i, y_i)\}_{i=1}^n$, OTD minimises
\begin{equation*}
\mathcal L_n(\theta,\psi)=\frac1n\sum_{i=1}^n S_\text{ES} \left( P^{\theta,\psi}(\cdot\mid x_i), y_i \right) + \lambda\mathcal A(\psi),
\end{equation*}
where $\lambda >0$ controls the strength of the regularisation.

This distinction separates OTD from approaches that first train a deterministic predictor and subsequently attach an uncertainty estimator. Here, stochasticity enters the predictive model itself and is optimised according to the quality of the induced conditional distribution.

\subsection{Predictive dispersion under model misspecification}
\paragraph{Misspecified predictive models.} Let $P_x^\ast$ denote the unknown conditional data-gene-rating law and define the family of distributions representable by OTD as $$\mathcal M = \left\{ P^{\theta,\psi}(\cdot \mid x): (\theta,\psi)\in\Theta\times\Psi \right\}.$$ In general, there is no reason to assume that $P^\ast\in\mathcal M$. The model may therefore be misspecified, in which case energy-score minimisation targets a best-in-class predictive law, rather than recovering  $P^\ast$ exactly. To characterise the target induced by the scoring rule, consider the unregularised population risk
$$
\mathcal R(\theta, \psi) = \mathbb E_X \mathbb E_{Y\sim P_X^*} S_\text{ES} \left(P^{\theta, \psi}(\cdot \mid X), Y\right), \quad (\theta^\dagger, \psi^\dagger) \in \arg \min_{\theta, \psi} \mathcal R(\theta, \psi).
$$
When $P^\ast\notin\mathcal M$, a population-risk minimiser $(\theta^\dagger,\psi^\dagger)$ defines a best-in-class predictive model within the OTD class. From this perspective, transporting the perturbation law enlarges the family of predictive distributions accessible to the predictive model relative to a fixed masking distribution. To assess the quality of the predictive model distribution, we employ both calibration and accuracy metrics, detailed in \autoref{sec:appendix_metrics}.

\paragraph{Deterministic targets.} The interpretation is particularly relevant for deterministic scientific models. If the high-fidelity map is $y= f(x)$, then the conditional data-generating law is $P_x^\ast=\delta_{f(x)}$. There is therefore no irreducible output variability to recover. Predictive dispersion should instead be interpreted as reflecting uncertainty associated with finite data and model-class limitations when learning $f:\mathcal X \to \mathcal Y$. In the idealised limit of complete information and correct specification, strict propriety of the Energy Score implies that the optimal predictive distribution collapses to the deterministic target. Consequently, in this case, the stochasticity introduced by OTD does not presuppose that the underlying physical system is stochastic. OTD uses a learnable distribution of model perturbations to induce predictive variability associated with finite data and model-class limitations. We therefore interpret the resulting dispersion as a candidate signal of epistemic uncertainty, and evaluate this interpretation explicitly through controlled experiments on data availability, model misspecification, and error--dispersion alignment.

\section{Results}
We first consider controlled settings in which specific factors affecting predictive uncertainty can be examined systematically. These experiments are designed to investigate three distinct properties of OTD: (i) whether transported masks enable capturing structured aleatoric uncertainty, (ii) whether predictive dispersion responds to structural model misspecification, and (iii) how this dispersion changes as model capacity increases. We then consider a functional regression problem, in which we assess the spatial alignment between predictive dispersion and prediction error, and whether dispersion decreases as data availability increases. These diagnostics provide empirical evidence for assessing its interpretation as a signal of reducible uncertainty.

\subsection{Controlled experiments}
\paragraph{Bimodal aleatoric experiment}

\begin{figure}[t]
    \centering
    \includegraphics[width=0.8\linewidth]{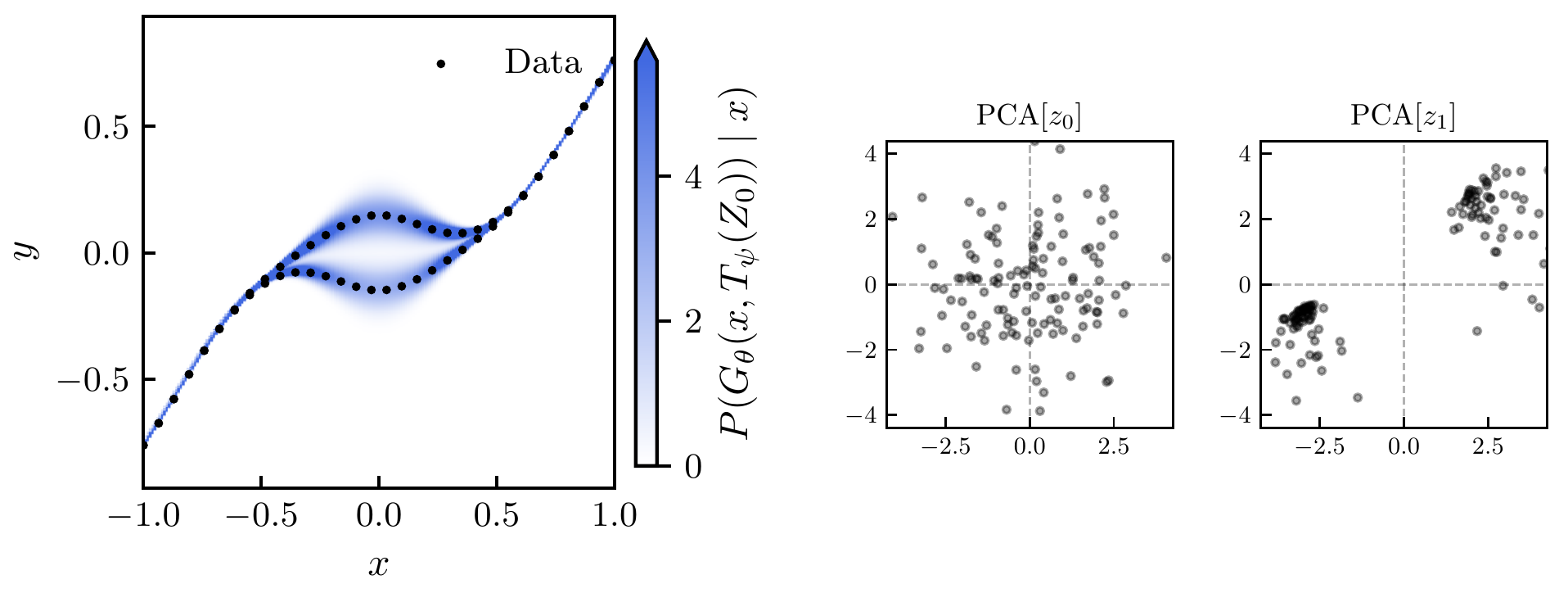}
    \caption{Bimodal aleatoric experiment. Left: kernel density estimation of the predictive conditional distribution. Right: principal component analysis of the first two components of the sampled and the transported masks.}
    \label{fig:bimodal}
\end{figure}

We first consider a controlled regression problem with a bimodal conditional target law whose two deterministic branches coincide over part of the input domain and bifurcate elsewhere (\autoref{fig:bimodal}, left). The experiment tests whether OTD collapses to a Dirac measure where the two branches coincide and recovers bimodality where the two separate. For the details, see \autoref{sec:appendix_experiments_bimodal}. Starting from an independent reference mask law, OTD learns a transported distribution whose push-forward through the predictive model reproduces this input-dependent transition. This isolates the expressiveness gained by transport: a single global latent law can induce qualitatively different predictive distributions through $z\mapsto G_\theta(x,z)$, and the bimodality is reflected also in the latent space (\autoref{fig:bimodal}, right).

\paragraph{Reduced-Order Model misspecification experiment}
We next consider a deterministic target map $f:\mathcal X \to \mathcal Y$ approximated within a restricted model class $\mathcal M_R$. When $f\notin\mathcal M_R$, a non-zero approximation error remains even after optimisation, providing a controlled setting for model misspecification. We refer to this as a reduced-order-model (ROM) misspecification experiment because it mimics a common ROM setting, where a restricted approximation space captures the solution's dominant components while leaving finer-scale features unresolved. For details, see \autoref{sec:appendix_experiments_rom}. The experiment tests whether OTD preserves predictive dispersion in the presence of this unresolved structural error. In this setting, OTD achieves lower $L^1$, $L^2$, Energy Score, and Mean Absolute Calibration Error (MACE) than MC Dropout (Figure \ref{fig:rom} and Table \ref{tab:otd_vs_dropout_metrics}). For all the metric definitions, see \autoref{sec:appendix_metrics}.

\begin{center}
    \begin{minipage}[t]{0.48\textwidth}
        \vspace{0pt}
        \centering

        \includegraphics[width=0.75\linewidth]{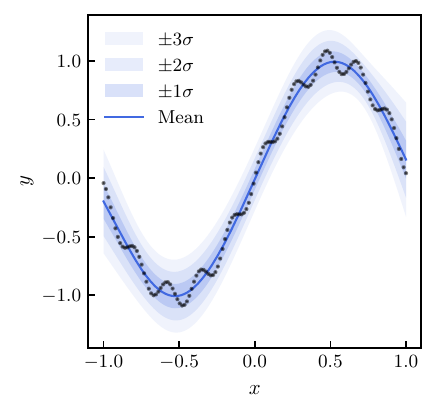}

        \captionof{figure}{
            Reduced-Order Model misspecification experiment. Model predictions, in blue, against data, dotted in black.}
        \label{fig:rom}
    \end{minipage}
    \hfill
    \begin{minipage}[t]{0.5\textwidth}
        \vspace{0pt}
        \centering
        \footnotesize

        \vspace{3em}

        \captionof{table}{Reduced-Order Model misspecification experiment. Comparison between OTD and MC Dropout on aggregate predictive metrics.}
        \label{tab:otd_vs_dropout_metrics}

        \vspace{0.5em}
        
        \begin{tabular}{ccc}
            \toprule
            Metric & OTD & MC Dropout \\
            \midrule
            $L^2$   & $\mathbf{0.072 \pm 0.001}$ & $0.092 \pm 0.001$ \\
            $L^1$   & $\mathbf{0.065 \pm 0.001}$ & $0.074 \pm 0.001$ \\
            $S_{\mathrm{ES}}$   & $\mathbf{0.042 \pm 0.001}$ & $0.055 \pm 0.001$ \\
            MACE & $\mathbf{0.077 \pm 0.025}$ & $0.183 \pm 0.010$ \\
            \bottomrule
        \end{tabular}
    \end{minipage}
\end{center}

\begin{figure}[ht]
    \includegraphics[width=0.95\linewidth]{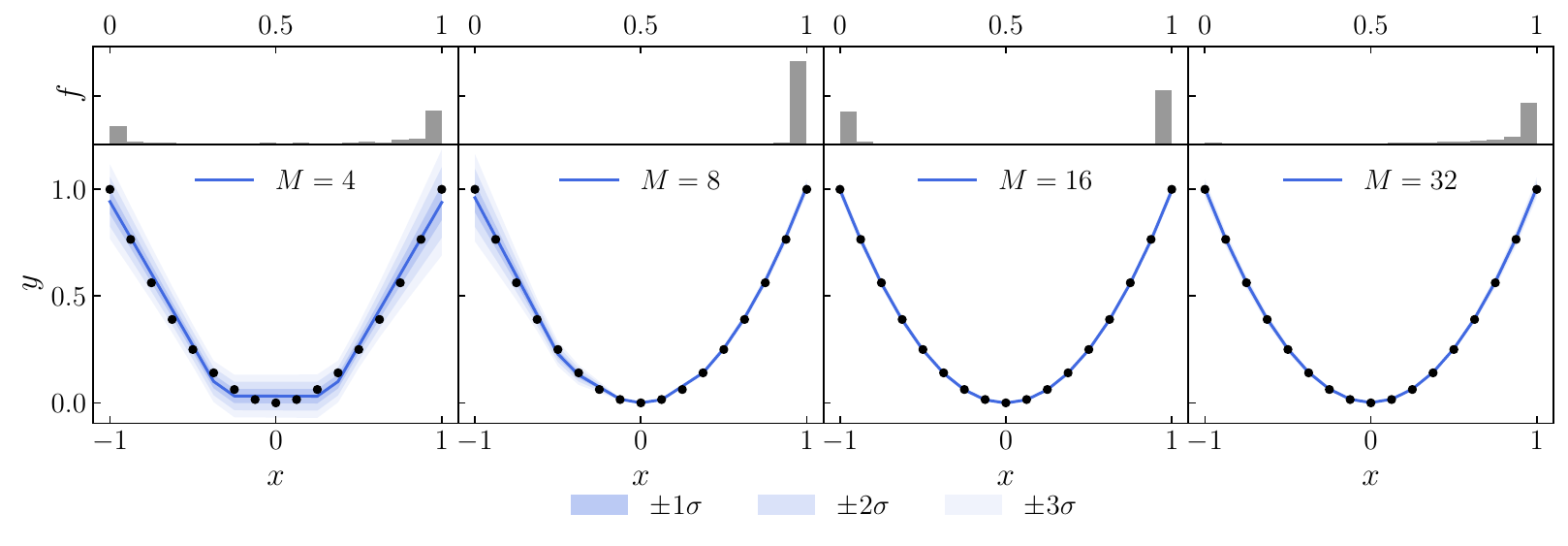}
    \caption{From under- to over-parametrisation experiment. The four images show the predictions of the trained models in blue with increasing hidden widths $M$. Above them, in grey, the histograms represent the distribution of the transported mask values across hidden units.}
    \label{fig:miss}
\end{figure}

\paragraph{From under- to over-parametrisation experiment} We next investigate how predictive dispersion changes with the predictive network expressivity at fixed sample size. We use a one-hidden-layer network, so that its representational capacity is controlled directly by the hidden width. We vary this width while keeping the dataset and optimisation protocol fixed. As the hidden width increases, we observe a contraction of predictive dispersion once the predictive model becomes sufficiently expressive (\autoref{fig:miss}). The experiment therefore provides a controlled transition from under- to over-parametrisation: stochasticity persists when the network lacks sufficient representational capacity and decreases once the target map can be represented more accurately. For the details, see \autoref{sec:appendix_experiments_collapse}.

Taken together, these experiments show that the transported mask distribution is not merely a source of generic output noise. Its geometry adapts to the predictive problem: it can represent multimodal predictive laws, remain dispersed under structural misspecification, and contract as the predictive model's capacity increases.

\subsection{Functional uncertainty quantification}

\paragraph{Ginzburg--Landau equilibrium fields experiment.} We consider functional regression of deterministic Ginzburg--Landau equilibrium fields, a canonical phase-field model describing the spatial and temporal evolution of a system undergoing a phase transition. For each parameter $\mu \in [0,1]$, OTD generates an ensemble of spatial predictions $(\mu,x,y)\mapsto u_\mu(x,y)$ in the unit square. Here, $u$ is a spatial order parameter describing the local state of a phase-transition system, while $\mu$ controls the local reaction dynamics and the spatial forcing. Predictive variability is evaluated as a candidate signal of epistemic uncertainty associated with finite training data and surrogate approximation. We evaluate predictive accuracy, marginal coverage, ensemble concentration, and spatial discrepancy--dispersion association. For details, see \autoref{sec:appendix_experiments_gl}.

\paragraph{Accuracy, marginal calibration, and sharpness.} OTD achieves better accuracy and probabilistic performance, with nearly nominal 90\% coverage and substantially sharper intervals, than Dropout (Table~\ref{tab:gl_metrics}).

\begin{table}[ht]
\centering
\footnotesize
\setlength{\tabcolsep}{3pt}
\caption{Ginzburg--Landau equilibrium fields experiment. Predictive performance and marginal calibration. Interval entries report PICP in per cent, with relative width in parentheses.}
\label{tab:gl_metrics}
\resizebox{\linewidth}{!}{%
\begin{tabular}{@{}l|cccc|ccccc@{}}
\toprule
& \multicolumn{4}{c|}{Predictive performance}
& \multicolumn{5}{c}{PICP \% (Width)} \\
\cmidrule(lr){2-5}
\cmidrule(lr){6-10}
Method & RMSE & MAE & $S_{\text{ES}}$ & MACE
& 50\% & 75\% & 80\% & 90\% & 95\% \\
\midrule
Deterministic
& 0.163 & 0.070 & 0.070 & --
& -- & -- & -- & -- & -- \\

Dropout
& 0.172 & 0.102 & 0.071 & 0.016
& \textbf{50.5} (0.068)
& \textbf{74.0} (0.115)
& 78.4 (0.128)
& 87.5 (0.164)
& 92.7 (0.193) \\

OTD
& \textbf{0.128}
& \textbf{0.044}
& \textbf{0.031}
& \textbf{0.009}
& 50.9 \textbf{(0.025)}
& 76.7 \textbf{(0.044)}
& \textbf{81.5 (0.050)}
& \textbf{90.5 (0.066)}
& \textbf{94.9 (0.081)} \\
\bottomrule
\end{tabular}
}
\end{table}

\paragraph{Ensemble dispersion versus data availability.} As the training-set size increases, the ensemble dispersion $\mathcal U(\hat P_{\mathcal D_n})$ monotonically decreases while predictive accuracy improves, where
$$
\mathcal U_{d_\mathcal Y}(P)
=
\frac12\int_{\mathcal X}
\mathbb E_{U,U'\sim P_x}d_\mathcal Y(U,U')\,\nu(\mathrm dx).
$$
This contraction indicates that OTD responds to the information available in the data, as expected of a measure of finite-data epistemic uncertainty.
\begin{center}
\footnotesize
\begin{tabular}{c|cccc}
     Training set size \% & 0.25 & 0.50 & 0.75 & 1.00 \\
     \midrule
     $\mathcal U(\hat P_{\mathcal D_n})$ & 0.16060 & 0.08232 & 0.05089 & 0.04967
\end{tabular}
\end{center}

\begin{figure}[h]
\centering
\includegraphics[width=0.6\linewidth]{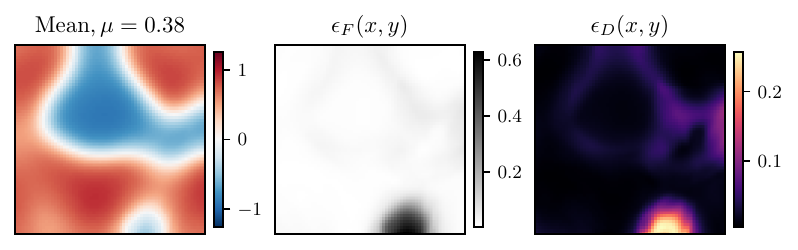}
\par\medskip
\includegraphics[width=0.6\linewidth]{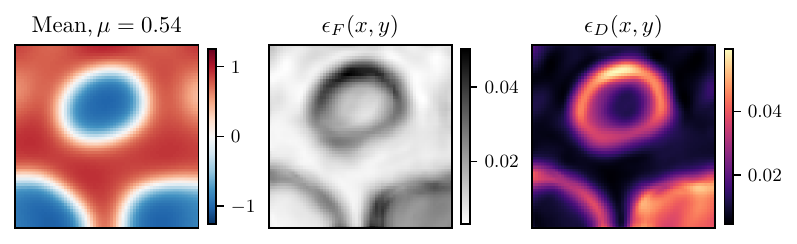}
\caption{Ginzburg--Landau equilibrium fields experiment. OTD ensemble mean, local predictive discrepancy, and pairwise dispersion for two selected test fields. These maps illustrate the spatial alignment between predictive discrepancy $\epsilon_F$ and predictive dispersion $\epsilon_D$.}
\label{fig:gl_predictions}
\end{figure}

\paragraph{Spatial discrepancy--dispersion association: beyond mean and variance.} We further assess whether predictive dispersion identifies where the surrogate is inaccurate, rather than merely matching aggregate error statistics. The strong rank association between spatial discrepancy and dispersion shows that OTD assigns greater uncertainty to regions with larger prediction errors, capturing spatially structured uncertainty beyond global mean and variance. See, for example, \autoref{fig:gl_predictions}. For quantitative details, refer instead to \autoref{sec:appendix_experiments_gl}

\paragraph{Disentangling the contributions of the Energy Score and transport flow.} We compare five controlled variants to separate the contributions of the Energy Score, transport flow, and kinetic regularisation. All OTD variants are trained with the Energy Score; OTD-0 denotes the zero-step model, which applies no transport and samples independent masks directly from the reference Concrete distribution. As shown in \autoref{tab:component_analysis}, training MC Dropout with the Energy Score moderately improves RMSE, MAE, and $S_{\mathrm{ES}}$, but worsens calibration. OTD-0 provides no further benefit, suggesting that the scoring rule and reference mask distribution alone do not explain the performance of OTD. Activating the flow substantially improves all metrics, reducing the RMSE and the $S_{\mathrm{ES}}$. Kinetic regularisation has a smaller effect: the unregularised model achieves the lowest RMSE, whereas complete OTD provides the best MAE, Energy Score, and MACE. Notably, complete OTD reduces RMSE by approximately $25\%$ relative to MC Dropout, despite not directly optimising it. Overall, the results identify the transport flow as the principal source of improvement. Further details, including sensitivity analyses for $\tau$ and the number of integration steps, are provided in \autoref{sec:appendix_experiments_gl}.

\begin{table}[h]
    \centering
    \caption{Ginzburg--Landau equilibrium fields: controlled comparison of the Energy Score, transport flow, and kinetic regularisation. Results are mean $\pm$ standard deviation over three seeds.}
    \label{tab:component_analysis}
    \resizebox{\linewidth}{!}{%
    \begin{tabular}{l|cccc}
        \toprule
        Model
        & RMSE
        & MAE
        & $S_{\mathrm{ES}}$
        & MACE \\
        \midrule
        MC Dropout
            & $0.1992 \pm 0.0158$
            & $0.1228 \pm 0.0088$
            & $0.0885 \pm 0.0075$
            & $0.0381 \pm 0.0045$ \\
        MC Dropout + ES
            & $0.1902 \pm 0.0133$
            & $0.1083 \pm 0.0067$
            & $0.0812 \pm 0.0056$
            & $0.1657 \pm 0.0106$ \\
        OTD-0
            & $0.2070 \pm 0.0077$
            & $0.1215 \pm 0.0019$
            & $0.0896 \pm 0.0012$
            & $0.1665 \pm 0.0018$ \\
        OTD, $\lambda_{\mathrm{kin}}=0$
            & $\mathbf{0.1442 \pm 0.0141}$
            & $0.0668 \pm 0.0037$
            & $0.0460 \pm 0.0028$
            & $0.0536 \pm 0.0273$ \\
        OTD
            & $0.1497 \pm 0.0152$
            & $\mathbf{0.0661 \pm 0.0084}$
            & $\mathbf{0.0455 \pm 0.0053}$
            & $\mathbf{0.0328 \pm 0.0225}$ \\
        \bottomrule
    \end{tabular}
    }
\end{table}

\subsection{Regression benchmarks on real-world data}
We finally evaluate OTD on a collection of standard tabular regression datasets from OpenML, adapted from Table 4 in \cite{franco_nonparametric_2025}. These experiments test whether the behaviour observed in the controlled and functional settings transfers to heterogeneous regression problems without relying on problem-specific structure, where aleatoric uncertainty also comes into play. All methods use the same deterministic backbone and data splits and, where applicable, follow the experimental protocols established in the literature \cite{franco_nonparametric_2025}. This allows differences in performance to be attributed primarily to the uncertainty mechanism. Dataset statistics, optimisation settings, and the complete evaluation protocol are reported in \autoref{sec:appendix_experiments_openml}.

\begin{figure}[h]
    \centering
    \includegraphics[width=0.45\linewidth]{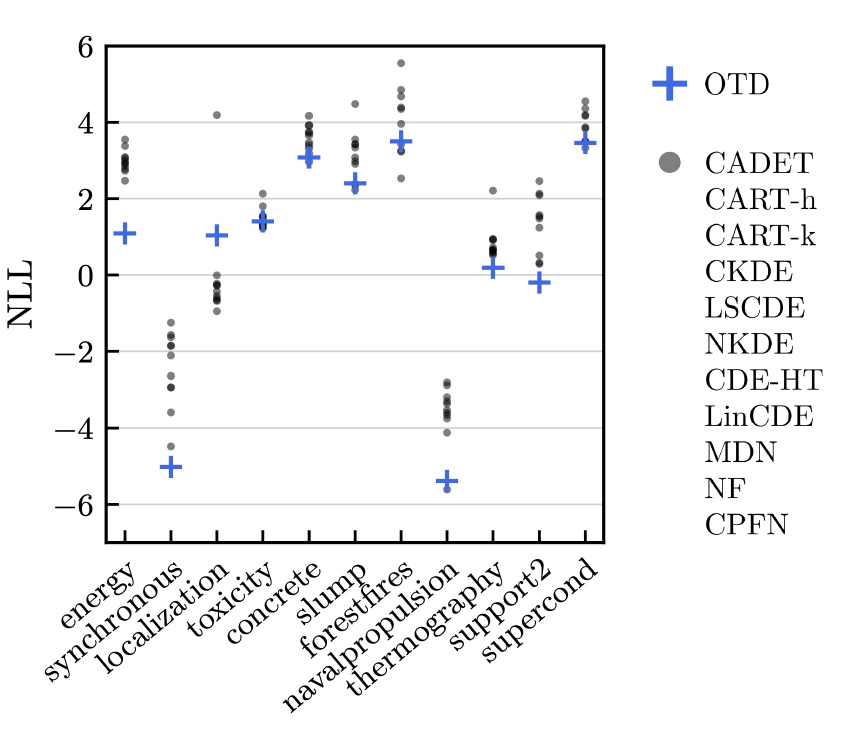}
    \caption{Experimental dataset. Negative log-likelihood comparison of OTD (blue) against other models (grey).}
    \label{fig:uci_predictive}
\end{figure}

Across the considered datasets, OTD remains competitive while improving the quality of the returned information, not only a point prediction, but a complete predictive distribution. The aggregate results in \autoref{fig:uci_predictive} indicate that learning the perturbation distribution can improve predictive quality without requiring an ensemble of independently trained networks. Subsection \ref{sec:appendix_experiments_openml} reports the complete numerical values and further investigations.

\section{Discussion and limitations}
OTD replaces the fixed, factorised perturbation law of standard Monte Carlo dropout with one learned through latent transport, enabling a single network to generate correlated, anisotropic, and potentially multimodal predictions. Controlled experiments show that OTD captures multimodal targets, responds to model misspecification, and contracts as data availability or model capacity increases. Functional regression results further show improved calibration and Energy Score without explicit Bayesian inference or independently trained ensembles.

Predictive dispersion must nevertheless be interpreted carefully. Although its persistence under stochastic targets, response to misspecification, contraction with data or capacity, and alignment with error support its interpretation as uncertainty, OTD does not explicitly separate aleatoric, model-class, and finite-data contributions. Nor does it define a Bayesian posterior: uncertainty arises from internal perturbations jointly learned with the predictor. Its validity has thus been assessed through controlled experimental settings.

OTD also adds computational cost through predictive sampling and integration of the latent flow, while the direct Monte Carlo Energy Score scales quadratically with the sample count. Nevertheless, our analysis in \autoref{sec:appendix_timing} shows stable performance with few samples and integration steps, and competitive results with low-dimensional transports in \autoref{sec:appendix_generalisation}. Validation on larger architectures, such as Fourier neural operators, and larger output spaces remains necessary and is currently under investigation.

Overall, OTD offers a simple way to enrich stochastic predictors beyond independent perturbations. Its combination of latent transport, proper scoring rules, and kinetic regularisation also suggests a variational connection with stochastic control and optimal transport, whose rigorous analysis is the subject of ongoing work.

\subsubsection*{Acknowledgments}
The present research has received support from the project FIS, MUR, Italy 2025-2028, Project code: FIS-2023-02228, CUP: D53C24005440001, ``SYNERGIZE: Synergizing Numerical Methods and Machine Learning for a new generation of computational models''.
The authors of this work are members of GNCS, ``Gruppo Nazionale per il Calcolo Scientifico'' (National Group for Scientific Computing) of INdAM (Istituto Nazionale di Alta Matematica).

\section*{AI use statement}
We used LLMs as assistive tools in two limited capacities: (i) retrieval and discovery, namely, to help identify potentially relevant related work, all of which was subsequently read and verified by the authors; and (ii) minor language editing to improve grammar and readability. All research ideas, theoretical results and proofs, and experiments were conceived, carried out, and verified independently by the authors. LLMs were not used to generate research ideas, derive results, conduct the experiments, or analyse data. The authors take full responsibility for the entire content of this paper.

\bibliographystyle{hieeetr}
\bibliography{bibliography}

\newpage

\appendix
\section{Diagnostic metrics}
\label{sec:appendix_metrics}
Throughout this work, we extensively use a set of diagnostic metrics to assess both the accuracy and the calibration of the predictive distribution. Let $\mathcal D_n = \{(x_i,y_i)\}_{i=1}^n$ denote the test set. For each input $x_i \in \mathcal X \subseteq \mathbb R^{d_x}$, let $\hat y_i^{(1)},\dots,\hat y_i^{(K)} \mathcal Y \subseteq \mathbb R^{d_y}$ be the samples from the predictive distribution $\hat P_{x_i}$. For any $i=1,\dots, n$, define the corresponding predictive mean
$$
\bar y_i = \frac{1}{K}\sum_{k=1}^K \hat y_i^{(k)}.
$$
The following metrics will be used.

\paragraph{Root Mean Squared Error (RMSE).}
We assess the accuracy of the predictive mean through
$$
\mathrm{RMSE}
=
\sqrt{\frac{1}{n}\sum_{i=1}^{n}
|y_i-\bar y_i|^2}.
$$
It measures the typical discrepancy between predictive means and observed targets, in the same units as the target. A small RMSE does not imply correctly calibrated predictive dispersion, and the squaring of errors makes this metric particularly sensitive to large discrepancies.

\paragraph{Mean Absolute Error (MAE).}
Another accuracy metric for the predictive mean is
$$
\mathrm{MAE}
=
\frac{1}{n}\sum_{i=1}^{n}
|y_i-\bar y_i|.
$$
It measures the average absolute discrepancy between the predictive means and the observed targets. Unlike the MSE, the MAE penalises errors linearly and is therefore less sensitive to large errors and outliers. Nevertheless, it only assesses the accuracy of the predictive mean: a small MAE does not imply that the predictive dispersion, calibration, or distributional shape is correctly represented.

\paragraph{Energy Score ($S_{\text{ES}}$).}
For each test point, the empirical Energy Score is defined as
$$
S_{\mathrm{ES}_i}
=
\frac{1}{K}\sum_{k=1}^{K}
\|\hat y_i^{(k)}-y_i\|_2
-
\frac{1}{2K(K-1)}
\sum_{\substack{k,r=1\\k\neq r}}^{K}
\|\hat y_i^{(k)}-\hat y_i^{(r)}\|_2.
$$
The first term measures the discrepancy between predictive samples and
the observation, whereas the second accounts for the internal dispersion
of the predictive distribution. Lower values indicate better predictive
performance. The reported aggregated scores are obtained by averaging over the test set:
$$
S_{\mathrm{ES}}
=
\frac{1}{n}\sum_{i=1}^{n}S_{\mathrm{ES}_i}.
$$

\paragraph{Predictive Interval Coverage Probability (PICP).}
For a nominal coverage level $\alpha\in(0,1)$, let
$$
I_i^{(\alpha)}
=
\left[
q_i\!\left(\frac{1-\alpha}{2}\right),
q_i\!\left(\frac{1+\alpha}{2}\right)
\right]
$$
denote the central empirical predictive interval, where $q_i(\tau)$ is the empirical $\tau$-quantile of the predictive samples $\{\hat y_i^{(k)}\}_{k=1}^K$

$$
q_i(\tau) = \inf\left\{ z\in\mathbb R: \hat F_i(z)\geq\tau \right\}, \qquad \hat F_i(z) = \frac1K\sum_{k=1}^{K} \mathbf 1_{\{\hat y_i^{(k)}\leq z\}}.
$$

The Predictive Interval Coverage Probability is defined as
$$
\mathrm{PICP}(\alpha)
=
\frac{1}{n}
\sum_{i=1}^n
\mathbf{1}\!\left\{y_i\in I_i^{(\alpha)}\right\}.
$$
A calibrated predictive distribution should satisfy $\mathrm{PICP}(\alpha)\approx\alpha$. Coverage alone, however, does not assess interval sharpness and should therefore be reported together with a measure of predictive interval width.

\paragraph{Sharpness.} We quantify interval sharpness by the mean width of the central predictive intervals,
$$
\mathrm{Sharpness}(\alpha)
=
\frac{1}{n}\sum_{i=1}^{n}
\left[
q_i\!\left(\frac{1+\alpha}{2}\right)
-
q_i\!\left(\frac{1-\alpha}{2}\right)
\right].
$$
Smaller values indicate narrower intervals, but are desirable only when accompanied by adequate coverage.

\paragraph{Mean Absolute Coverage Error (MACE).}
Let $\mathcal A=\{\alpha_1,\dots,\alpha_L\}\subset(0,1)$ be a finite set of nominal coverage levels. The Mean Absolute Coverage Error is defined as
$$
\mathrm{MACE}
=
\frac{1}{L}
\sum_{\ell=1}^{L}
\left|
\mathrm{PICP}(\alpha_\ell)-\alpha_\ell
\right|.
$$
It measures the average absolute discrepancy between empirical and nominal coverage across the considered levels. A perfectly calibrated predictive distribution would attain $\mathrm{MACE}=0$, while lower values indicate better marginal calibration.

Taken together, these metrics highlight different facets of how well the predictive distribution performs, and they should be interpreted jointly.

\section{Energy-Score Optimisation}
\label{sec:appendix_es}

\paragraph{Restricted energy-distance projection.} Consider the case in which the target is deterministic, namely, it is not affected by aleatoric uncertainty. Let $P\in\mathcal P_1(\mathbb R)$ and $y\in\mathbb R$. The Energy Score associated with $d_{\mathcal Y}(u,v)=|u-v|$ is
$$
S_{\mathrm{ES}}(P,y)
=
\mathbb E|Y-y|
-
\frac12\mathbb E|Y-Y'|,
\qquad
Y,Y'\overset{\mathrm{i.i.d.}}{\sim}P.
$$
Since $\mathcal E_{| \cdot |}(P,\delta_y)=2\mathbb E|Y-y|-\mathbb E|Y-Y'|$, we have
$$
S_{\mathrm{ES}}(P,y) = \frac12\mathcal E_{|\cdot |}(P,\delta_y).
$$
Consequently, over $\mathcal P_1(\mathbb R)$, the unique minimiser is the deterministic law $\delta_y$. Under the unregularised Energy Score, positive predictive dispersion can therefore be optimal only when the admissible predictive family does not contain $\delta_y$.

\subsubsection*{Gaussian Scale Family.}

\paragraph{Proposition.} Consider $Y_s\sim\mathcal N(\mu,s^2),\ s\geq 0$. If $\mu=y$, then $S_{\text{ES}}(\mathcal N(\mu,s^2),y)$ is uniquely minimised at $s=0$. If $\mu\neq y$, it admits the unique minimiser at
$$
s_\ast
=
\frac{|\mu - y|}{\sqrt{\log 2}}
\simeq
1.2011\,|\mu - y|.
$$

Thus, even for a deterministic target, minimising the Energy Score can favour non-zero predictive dispersion when the predictive mean is inaccurate. In OTD, this means that dispersion may reflect systematic surrogate error as well as uncertainty arising from limited data; it should not, by itself, be interpreted as a separate estimate of either source.

\textit{Proof.} Set $r=|\mu-y|$. Since $Y_s-Y_s'\sim\mathcal N(0,2s^2),$ we have
$$
\frac12\mathbb E|Y_s-Y_s'|
=
\frac{s}{\sqrt{\pi}}.
$$
Moreover, for $s>0$, the folded-normal identity gives
$$
\mathbb E|Y_s-y|
=
s\left[
2\varphi\left(\frac{r}{s}\right)
+
\frac{r}{s}
\left(
2\Phi\left(\frac{r}{s}\right)-1
\right)
\right],
$$
where $\varphi$ and $\Phi$ denote the standard Gaussian density and distribution function. Therefore,
$$
S_{\text{ES}}(\mathcal N(\mu,s^2),y)
=
s\left[
2\varphi\left(\frac{r}{s}\right)
+
\frac{r}{s}
\left(
2\Phi\left(\frac{r}{s}\right)-1
\right)
-
\frac1{\sqrt{\pi}}
\right].
$$
If $r=0$, then
$$
S_{\text{ES}}(\mathcal N(\mu,s^2),y)
=
s\left(
\sqrt{\frac{2}{\pi}}
-
\frac1{\sqrt{\pi}}
\right),
$$
which is strictly increasing and hence uniquely minimised at $s=0$.

Assume $r>0$ and set $a=r/s$. Since
$$
S_{\text{ES}}(\mathcal N(\mu,s^2),y)
=
rF(a),
\qquad
F(a)
=
\frac{
2\varphi(a)
+
a\bigl(2\Phi(a)-1\bigr)
-
1/\sqrt{\pi}
}{a},
$$
minimising over $s>0$ is equivalent to minimising $F$ over $a>0$. Differentiation yields
$$
F'(a)
=
\frac{1/\sqrt{\pi}-2\varphi(a)}{a^2}.
$$
Thus $F'(a)=0$ if and only if
$$
2\varphi(a)=\frac1{\sqrt{\pi}},
$$
or equivalently $a^2=\log 2$. Since $\varphi$ is strictly decreasing on $(0,\infty)$, $F'$ changes sign exactly once, from negative to positive. Hence the minimiser is unique and
$$
s_\ast
=
\frac{r}{a_\ast}
=
\frac{|\mu-y|}{\sqrt{\log 2}}.
$$
\hfill$\square$

\subsubsection*{General Centred Location--Scale Families.}
The preceding mechanism is not specific to Gaussian distributions. Let $X$ be a non-degenerate real-valued random variable satisfying $\mathbb EX=0, \ \mathbb E|X|<\infty$, and consider
$$
Y_s=\mu+sX,
\qquad
P_s=\mathcal L(Y_s),
\qquad
s\geq0.
$$
Writing $e=\mu-y$ and letting $X'$ be an independent copy of $X$, the Energy Score becomes
$$
S_{\mathrm{ES}}(P_s,y)
=
\mathbb E|e+sX|
-
\frac{s}{2}\mathbb E|X-X'|.
$$

\paragraph{Proposition.} The function $s\mapsto S_{\mathrm{ES}}(P_s,y)$ is convex and continuous on $[0,\infty)$. If $e=0$, its unique minimiser is $s=0$. If $e\neq0$, it admits at least one global minimiser, and every global minimiser satisfies $s_\ast>0$.

\textit{Proof.} Convexity follows because $s\mapsto|e+sX|$ is convex for every realisation of $X$, whereas the second term is linear in $s$. Continuity follows from
$$
\bigl||e+sX|-|e+tX|\bigr|
\leq
|s-t|\,|X|.
$$

If $e=0$, then
$$
S_{\mathrm{ES}}(P_s,y)
=
s\left(
\mathbb E|X|
-
\frac12\mathbb E|X-X'|
\right).
$$
The coefficient is strictly positive because $2\mathbb E|X|-\mathbb E|X-X'|=\mathcal E(\mathcal L(X),\delta_0)>0$ for every non-degenerate $X$. Hence $S_{\mathrm{ES}}(P_s,y)>S_{\mathrm{ES}}(P_0,y)$ for every $s>0$. Suppose now that $e\neq0$. The bound
$$
\left|
\frac{|e+sX|-|e|}{s}
\right|
\leq |X|
$$
allows differentiation at $s=0^+$ by dominated convergence, giving
$$
\left.
\frac{\mathrm d}{\mathrm ds}
\mathbb E|e+sX|
\right|_{s=0^+}
=
\operatorname{sign}(e)\mathbb E[X]
=
0.
$$
It follows that
$$
\left.\frac{\mathrm d}{\mathrm ds}S_{\mathrm{ES}}(P_s,y)\right|_{s=0^+}
=
-\frac12\mathbb E|X-X'|
<0.
$$
Therefore $s=0$ is not a local minimiser. Finally,
$$
\frac{S_{\mathrm{ES}}(P_s,y)}{s}
=
\mathbb E\left|X+\frac es\right|
-
\frac12\mathbb E|X-X'|
$$
converges as $s\to\infty$ to
$$
\mathbb E|X|
-
\frac12\mathbb E|X-X'|
>0.
$$
Thus $S_{\mathrm{ES}}(P_s,y)\to+\infty$ as $s\to\infty$. Continuity and coercivity guarantee the existence of a global minimiser, while the negative right derivative at the origin implies that every such minimiser is strictly positive.
\hfill$\square$

\subsubsection*{Other preliminary results}
\paragraph{Scaling with the residual.}
For $e\neq0$, writing $s=|e|t$ gives
$$
S_{\mathrm{ES}}(P_{|e|t},y)
=
|e|
\left[
\mathbb E
\left|
\operatorname{sign}(e)+tX
\right|
-
\frac{t}{2}\mathbb E|X-X'|
\right].
$$
Hence the optimal scale is proportional to the magnitude of the residual:
$$
s_\ast(e)
=
|e|\,t_\ast^{\operatorname{sign}(e)}.
$$
If $X\overset{d}{=}-X$, the coefficient is independent of the sign of $e$, and therefore $s_\ast(e)=c_X|e|.$ The Gaussian result corresponds to
$$
c_X=\frac1{\sqrt{\log 2}}.
$$
Whenever $\mathbb P(e+s_\ast X=0)=0$, an interior minimiser satisfies the first-order condition
$$
\mathbb E\left[
X\operatorname{sign}(e+s_\ast X)
\right]
=
\frac12\mathbb E|X-X'|.
$$

\paragraph{Symmetric two-point predictive family.}
Positive dispersion may also be optimal in a predictive family that is not described by a continuous scale distribution. Consider
$$
P_a
=
\frac12\delta_{\mu-a}
+
\frac12\delta_{\mu+a},
\qquad
a\geq0,
$$
and let $y=\mu+e$. Direct calculation gives
$$
S_{\mathrm{ES}}(P_a,y)
=
\frac12\left(|e-a|+|e+a|\right)-\frac a2
=
\max\{|e|,a\}-\frac a2.
$$
For $0\leq a\leq |e|$, this expression equals $|e|-a/2$ and is strictly decreasing, whereas for $a\geq|e|$ it equals $a/2$ and is strictly increasing. Its unique minimiser is therefore $a_\ast=|e|,$ with
$$
S_{\mathrm{ES}}(P_{a_\ast},y)
=
\frac{|e|}{2}.
$$
By comparison, the deterministic law $P_0=\delta_\mu$ gives $S_{\mathrm{ES}}(P_0,y)=|e|$. Thus, within this constrained symmetric family, the Energy Score prefers a bimodal predictive law to the misspecified deterministic prediction.

\paragraph{A mean constraint alone is insufficient.}
The previous conclusion depends essentially on the chosen predictive family. A fixed predictive mean alone does not yield a well-posed optimisation problem. Without loss of generality, let $y=0$ and require the predictive mean to equal $e\neq0$. For $p\in(0,1)$, define $P_p=(1-p)\delta_0+p\delta_{e/p}.$ Then $\mathbb E_{P_p}[Y]=e,$ while $\mathbb E_{P_p}|Y|=|e|$ and
$$
\frac12\mathbb E_{P_p}|Y-Y'|
=
(1-p)|e|.
$$
Consequently,
$$
S_{\mathrm{ES}}(P_p,0)
=
p|e|
\to 0
\qquad\text{as }p\downarrow0.
$$
It follows that
$$
\inf_{\substack{P\in\mathcal P_1(\mathbb R)\\
\mathbb E_P[Y]=e}}
S_{\mathrm{ES}}(P,0)
=
0,
$$
although the infimum is not attained when $e\neq0$. The predictive law can place almost all its mass at the observation while sending a vanishing amount of mass arbitrarily far away to preserve the constrained mean. Additional restrictions on the predictive family, its moments, its support, or its transport cost are therefore necessary.

\paragraph{Implications for transported predictive models.}
These results do not imply that the Energy Score intrinsically favours dispersion or multimodality: over an unrestricted class, its unique optimum for a deterministic observation is the Dirac measure concentrated at that observation. Positive dispersion arises when optimisation is restricted to a predictive family that cannot attain this optimum. Within such a family, dispersion can reduce the Energy Distance from the predictive law to the observation and thereby absorb part of the residual error. Predictive spread may therefore reflect model-class constraints in addition to aleatoric uncertainty. In OTD, the transported mask family determines the admissible predictive laws, while the kinetic regularisation discourages excessively distorted latent distributions. Since the transport and predictive network are jointly optimised, this regularisation does not by itself impose direct control in output space, but it provides a principled mechanism for balancing predictive fit against unnecessary deformation of the reference perturbation law.

\section{OTD Implementation Details}
\label{sec:appendix_implementation}

In this section, the details of the OTD method are thoroughly explained.

\paragraph{Relaxed mask sampling.} OTD starts from sampling a continuous relaxation of an independent Bernoulli mask. For each Monte Carlo realisation $k \in \{1,2,\dots, K\}$, we sample $U^{(k)} \sim \mathcal U(0,1)^{d_z}$, and define the initial logit mask vector
$$
m_0^{(k)}
=
\frac{
\operatorname{logit}(p)
+
\operatorname{logit}(U^{(k)})
}{\tau},
$$
where $\operatorname{logit}(x) =\log x-\log(1-x)$, $\tau>0$ is the relaxation temperature, and $p\in(0,1)$ is the probability of retaining each feature. The sum should be interpreted component-wise. The corresponding relaxed mask is
$$
z_0^{(k)}
=
\sigma\left(m_0^{(k)}\right)
\in(0,1)^{d_z},
$$
where $\sigma(\cdot)$ is the sigmoidal function. Small values of $\tau$ approach binary Bernoulli masks, whereas larger values yield smoother perturbations. In particular, for $p=1/2$ and $\tau=1$, the relaxed mask is uniform on $(0,1)$.

\paragraph{Logit-space transport.} The transport is performed in unconstrained logit space. For every Monte Carlo sample $k \in \{0, 1, \dots, K\}$,
$$
\frac{\mathrm d m_t^{(k)}}{\mathrm dt}
=
\tilde v_\psi
\left(
t,
\sigma(m_t^{(k)})
\right),
\qquad
m_{t=0}^{(k)}=m_0^{(k)},
$$
where $\tilde v_\psi: [0,1] \times \mathcal Z \to \mathbb R^{d_z}$ is the velocity field acting on the unconstrained masks, parametrised by a fully connected neural network. After integration, the transported masks are obtained as
$$
z_1^{(k)}
=
\sigma\left(m_1^{(k)}\right).
$$
The dynamics therefore evolve on $\mathbb R^{d_z}$, since there is no guarantee that the dynamics remain constrained in the unit cube, whereas the predictive network receives bounded masks in $(0,1)^{d_z}$.

Then, for instance, when OTD is applied to a fully connected architecture, the transported masks are partitioned according to the hidden layers $H \in \mathbb N$, each with width $d_i$,
$$
z_1^{(k)}
=
\left(
z_1^{(k,1)},\ldots,z_1^{(k,H)}
\right), \qquad z_1^{(k,i)} \in (0,1)^{d_i}, \ \forall i = 1,2,\dots, H,
$$
and multiplicatively modulate the corresponding hidden representations
$$
h_0^{(k)}=x,
\qquad 
h_i^{(k)}
=
\phi
\left(
W_i h_{i-1}^{(k)}+b_i
\right)
\odot z_1^{(k,i)},
$$
where $\phi$ is the scalar non-linear activation function acting on each vector element pointwisely, $W \in~\mathbb R^{d_i \times d_{i-1}}$ the weight matrices, and $b_i \in \mathbb R^{d_i}$ the bias vectors for all $i=1,2,\dots, H$.

We denote the combined actions of all the hidden layers as $G_\theta: \mathcal X \times \mathcal Z \to \mathcal Y$, where $\theta$ denotes the parameters of the chosen neural architecture. The final predictions are thus
$$
\hat y^{(k)}
=
G_\theta
\left(
x,z_1^{(k)}
\right).
$$

A single mask realisation is shared across the inputs associated with the same Monte Carlo draw. The output law nevertheless remains input-dependent because the transported mask acts on hidden representations that depend on $x$.

\paragraph{Euler and Heun integration.} The interval $[0,1]$ is discretised on a uniform grid
$$
0=t_0<t_1<\cdots<t_L=1,
\qquad
\Delta t=\frac1L.
$$
\begin{itemize}
\item For Euler integration,
$$
m_{\ell+1}^{(k)}
=
m_\ell^{(k)}
+
\Delta t\,
\tilde v_\psi
\left(
t_\ell,
\sigma(m_\ell^{(k)})
\right).
$$
\item For Heun integration, the predictor is
$$
\widetilde m_{\ell+1}^{(k)}
=
m_\ell^{(k)}
+
\Delta t\,
\tilde v_\psi
\left(
t_\ell,
\sigma(m_\ell^{(k)})
\right),
$$
and the corrected update is
$$
m_{\ell+1}^{(k)}
=
m_\ell^{(k)}
+
\frac{\Delta t}{2}
\left[
\tilde v_\psi
\left(
t_\ell,\sigma(m_\ell^{(k)})
\right)
+
\tilde v_\psi
\left(
t_{\ell+1},
\sigma(\widetilde m_{\ell+1}^{(k)})
\right)
\right].
$$
\end{itemize}

\paragraph{Monte Carlo Energy Score.} Given $K\in \mathbb N$ independent transported masks, the empirical predictive distribution is
$$
P_{\theta,\psi}^K(\cdot\mid x)
=
\frac1K
\sum_{k=1}^K
\delta_{G_\theta\left(x,z_1^{(k)}\right)}.
$$
For a single observation $(x,y)$, the Energy Score is approximated by
$$
\hat S_{\mathrm{ES}}^{K}
=
\frac1K
\sum_{k=1}^K
d_{\mathcal Y}
\left(
\hat y^{(k)},y
\right)
-
\frac{1}{2K(K-1)}
\sum_{k\neq j}
d_{\mathcal Y}
\left(
\hat y^{(k)},\hat y^{(j)}
\right).
$$

The first term measures agreement between predictive realisations and the observation, whereas the second measures internal predictive dispersion. At inference time, the same sampling procedure yields empirical means, variances, quantiles, and prediction intervals.

\paragraph{Kinetic--action approximation.} Since $z_t=\sigma(m_t)$, the induced velocity in mask space is
$$
\frac{\mathrm dz_t}{\mathrm dt}
=
z_t\odot(1-z_t)
\odot
\tilde v_\psi(t,z_t),
$$
preventing the dynamics from exiting the unit cube. The kinetic action is therefore evaluated in the bounded mask space as
$$
\mathcal A(\psi)
=
\frac12
\int_0^1
\int_\mathcal Z
\left\|
z_t\odot(1-z_t)
\odot
\tilde v_\psi(t,z_t)
\right\|_2^2
\, \mathrm dz\,\mathrm dt .
$$

For Euler integration, for $K\in\mathbb N$, its Monte Carlo approximation is
$$
\hat{\mathcal A}_{K}(\psi)
=
\frac1K
\sum_{k=1}^K
\frac12
\sum_{\ell=0}^{L-1}
\Delta t
\left\|
z_\ell^{(k)}
\odot
\left(1-z_\ell^{(k)}\right)
\odot
\tilde v_\psi
\left(
t_\ell,z_\ell^{(k)}
\right)
\right\|_2^2.
$$

\paragraph{Empirical training objective.} Given $K_{\text{ES}}, K_{\text{kin}} \in \mathbb N$, and $\lambda_{\text{kin}} > 0$, the complete empirical training objective reads
$$
\mathcal L(\theta,\psi)
=
\frac1n
\sum_{i=1}^n
\hat S_{\mathrm{ES}}^{K_{\text{ES}}}
\left(
P_{\theta,\psi}(\cdot\mid x_i),y_i
\right)
+
\lambda_{\mathrm{kin}}
\hat{\mathcal A}_{K_{\text{kin}}}(\psi).
$$
The Energy Score and kinetic action are estimated using independent Monte Carlo samples.

\section{Experimental Details}
\label{sec:appendix_experiments}

\subsection{Bimodal Dataset}
\label{sec:appendix_experiments_bimodal}

The bimodal experiment is designed as a controlled test of distributional expressiveness. The target observations admit two distinct predictive behaviours in a neighbourhood of $x=0$, while an almost deterministic function expression holds elsewhere (\autoref{fig:bimodal}). The experiment tests whether OTD can transform an initially unstructured reference mask law into a latent distribution whose push-forward through the predictive network becomes locally bimodal in output space.

The predictive distribution is analysed through three complementary diagnostics: the stochastic predictions themselves, a kernel-density estimate of the induced output distribution, and the correlation structure of the transported masks. A principal-component projection of the masks further shows that local bimodality in output space is associated with non-trivial global geometry in latent space. The dependence on the input is introduced by the predictive network $G_\theta(x,z)$, although the transported mask distribution itself is shared across inputs.

\autoref{tab:bimodal_hyperparameters} collects the hyperparameters required to replicate the experiment with the bimodal dataset. The target function is the following:
$$
y_\pm (x) = \tanh \left(x^3 \pm a e^{- wx^2}\right), \qquad x \in [-1,1],
$$
from which 32 points are drawn equally spaced for training from each branch of $y_\pm$, with split amplitude $a=0.15$ and split width $w=12$. The conditional law thus reads
$$
P_x^* = \frac12\delta_{y_+(x)} + \frac12\delta_{y_-(x)}, \qquad x \in [-1,1].
$$
The two branches are well separated around $x=0$ and become nearly indistinguishable away from this region.

\begin{figure}[ht]
    \centering
    \includegraphics[width=0.3\linewidth]{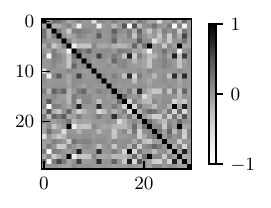}
    \caption{Bimodal aleatoric experiment. Empirical correlation of the latent variables transported by the flow in the bimodal dataset experiment.}
    \label{fig:corr}
\end{figure}

It is worth noticing that the mask produced by the latent flow presents a non-trivial correlation pattern (see \autoref{fig:corr}).

\begin{table}[t]
\caption{Bimodal aleatoric experiment. List of hyperparameters.}
\label{tab:bimodal_hyperparameters}
\footnotesize
\centering
\begin{minipage}[t]{0.48\linewidth}
\centering
\begin{tabular}{@{}ll@{}}
\toprule
\textbf{Hyperparameter} & \textbf{Value} \\
\midrule

\multicolumn{2}{@{}l}{\textbf{Latent mask}} \\
Concrete parameter $p$ & $0.5$ \\
Concrete temperature $\tau$ & $1.0$ \\

\addlinespace
\multicolumn{2}{@{}l}{\textbf{Latent flow}} \\
Velocity hidden widths & $(4,4)$ \\
Activation & GeLU \\
Kernel initialisation & He normal \\
Bias initialisation & Zero \\
Integration method & Heun \\
Integration steps & $80$ \\

\addlinespace
\multicolumn{2}{@{}l}{\textbf{Predictive Network}} \\
Widths & $(1,8,8,1)$ \\
Activation & GeLU \\
Kernel initialisation & He normal \\
Bias initialisation & Zero \\

\bottomrule
\end{tabular}
\end{minipage}
\begin{minipage}[t]{0.48\linewidth}
\centering
\begin{tabular}{@{}ll@{}}
\toprule
\textbf{Hyperparameter} & \textbf{Value} \\
\midrule

\multicolumn{2}{@{}l}{\textbf{Optimisation}} \\
Optimiser & AdamW \\
Learning rate & $10^{-3}$ \\
Weight decay & $10^{-5}$ \\

\addlinespace
\multicolumn{2}{@{}l}{\textbf{Training}} \\
Number of epochs & $50\,000$ \\

\addlinespace
\multicolumn{2}{@{}l}{\textbf{Loss}} \\
Energy-score MC samples & $32$ \\
Kinetic-action MC samples & $1$ \\
Kinetic regularisation & $10^{-4}$ \\

\bottomrule
\end{tabular}
\end{minipage}
\end{table}

\subsection{Reduced-Order Model misspecification Dataset}
\label{sec:appendix_experiments_rom}

The reduced-order experiment mimics a setting in which the high-fidelity response contains structures that cannot be represented exactly by a restricted predictive model. The deterministic target is generated from a sinusoidal function containing two oscillatory frequencies, while the predictive network is deliberately kept small.

The low-frequency component can be represented by the predictive model's mean, whereas the higher-frequency component induces a systematic approximation defect. The experiment therefore provides a controlled test of model-class misspecification: predictive dispersion can be compared with the unresolved component of the target rather than with observational noise.

Performance is assessed through point-prediction errors, Energy Score, and calibration error, and compared with standard dropout.

We consider the deterministic function (see \autoref{fig:bimodal})
$$
y(x)=\sin(\omega x)+\frac 1 {10}\sin(10\omega x), \qquad x\in[-1,1].
$$
The training set contains $128$ uniformly spaced input-output pairs, while evaluation uses $512$ uniformly spaced points, with frequency parameter $\omega =3$. Predictive uncertainty is estimated from $256$ stochastic model realisations. \autoref{tab:rom_hyperparameters} summarises the hyperparameters used for the experiment.

\begin{table}[ht]
\caption{Reduced-Order Model misspecification experiment. List of hyperparameters.}
\label{tab:rom_hyperparameters}
\footnotesize
\centering
\begin{minipage}[t]{0.48\linewidth}
\centering
\begin{tabular}{@{}ll@{}}
\toprule
\textbf{Hyperparameter} & \textbf{Value} \\
\midrule

\multicolumn{2}{@{}l}{\textbf{Latent mask}} \\
Concrete parameter $p$ & $0.5$ \\
Concrete temperature $\tau$ & $1.0$ \\

\addlinespace
\multicolumn{2}{@{}l}{\textbf{Latent flow}} \\
Velocity hidden widths & $(2,2)$ \\
Activation & GeLU \\
Kernel initialisation & He normal \\
Bias initialisation & Zero \\
Integration method & Euler \\
Integration steps & $80$ \\

\addlinespace
\multicolumn{2}{@{}l}{\textbf{Predictive Network}} \\
Widths & $(1,8,8,1)$ \\
Activation & GeLU \\
Kernel initialisation & He normal \\
Bias initialisation & Zero \\

\bottomrule
\end{tabular}
\end{minipage}
\begin{minipage}[t]{0.48\linewidth}
\centering
\begin{tabular}{@{}ll@{}}
\toprule
\textbf{Hyperparameter} & \textbf{Value} \\
\midrule

\multicolumn{2}{@{}l}{\textbf{Optimisation}} \\
OTD optimiser & AdamW \\
Dropout optimiser & Adam \\
Learning rate & $10^{-4}$ \\
Weight decay & $10^{-5}$ \\

\addlinespace
\multicolumn{2}{@{}l}{\textbf{Training}} \\
Number of epochs & $50\,000$ \\
Checkpoint interval & $5\,000$ \\

\addlinespace
\multicolumn{2}{@{}l}{\textbf{Loss}} \\
Energy-score MC samples & $64$ \\
Kinetic-action MC samples & $1$ \\
Kinetic regularisation & $10^{-5}$ \\
Dropout parameter regularisation & $10^{-3}$ \\

\addlinespace
\multicolumn{2}{@{}l}{\textbf{Dropout baseline}} \\
Widths & $(1,8,8,1)$ \\
Activation & GeLU \\
Dropout rate & $0.05$ \\

\bottomrule
\end{tabular}
\end{minipage}
\end{table}

\subsection{Latent collapse experiment}
\label{sec:appendix_experiments_collapse}

To investigate the transition from under- to over-parametrisation, we consider a shallow predictive network with varying hidden width $M\in\{4,8,16,32\}$. The deterministic teacher is a piecewise-linear, parabolic-like function with $M^\ast = 16$ linear segments.

For $M<M^\ast$, the predictive network is structurally unable to reproduce the target exactly and stochastic predictive variability may compensate for the unresolved component. Around $M=M^\ast$, a deterministic representation becomes possible in principle. In the over-parametrised regime, the predictive network can reduce its sensitivity to the transported mask, for example by relying on only a subset of the available hidden units. The experiment therefore examines how both output dispersion and the transported mask distribution change as representational capacity increases.

We consider the noiseless regression problem using $N=17$ uniformly spaced training points,
$$
y_i = x_i^2,\qquad x_i=-1+\frac{2i}{16},\qquad i=0,\ldots,16.
$$
The predictor is an OT-dropout MLP with one hidden ReLU layer of width ranging in $M \in [4, 8, 16, 32]$. Since a finite ReLU network represents a piecewise-linear function, it cannot reproduce the quadratic target exactly over the entire continuous domain, thereby introducing structural model misspecification (\autoref{fig:miss}). The model is trained in full-batch mode by minimising the Energy Score with kinetic regularisation. Predictive uncertainty is subsequently evaluated from $256$ stochastic realisations both at the training points and on a uniform grid of $1025$ points over $[-1,1]$. Refer to \autoref{tab:misspecification-hyperparameters} for the experiment's hyperparameters.

\begin{table}[ht]
\caption{From under- to over-parametrisation experiment. List of hyperparameters.}
\label{tab:misspecification-hyperparameters}

\footnotesize
\centering

\begin{minipage}[t]{0.48\linewidth}
\centering
\begin{tabular}{@{}ll@{}}
\toprule
\textbf{Hyperparameter} & \textbf{Value} \\
\midrule

\multicolumn{2}{@{}l}{\textbf{Latent mask}} \\
Mask dimension & $4, 8, 16, 32$ \\
Concrete parameter $p$ & $0.5$ \\
Concrete temperature $\tau$ & $1.0$ \\

\addlinespace
\multicolumn{2}{@{}l}{\textbf{Latent flow}} \\
Velocity hidden widths & $(4)$ \\
Velocity activation & ReLU \\
Velocity kernel initialisation & He uniform \\
Integration method & Heun \\
Integration steps & $60$ \\

\addlinespace
\multicolumn{2}{@{}l}{\textbf{Predictive Network}} \\
Widths & $(1,32,1)$ \\
Activation & ReLU \\
Kernel initialisation & He uniform \\

\bottomrule
\end{tabular}
\end{minipage}
\hfill
\begin{minipage}[t]{0.48\linewidth}
\centering
\begin{tabular}{@{}ll@{}}
\toprule
\textbf{Hyperparameter} & \textbf{Value} \\
\midrule

\multicolumn{2}{@{}l}{\textbf{Optimisation}} \\
Optimiser & AdamW \\
Learning rate & $10^{-2}$ \\
Weight decay & $10^{-5}$ \\

\addlinespace
\multicolumn{2}{@{}l}{\textbf{Training}} \\
Training scheme & Full batch \\
Number of epochs & $20\,000$ \\

\addlinespace
\multicolumn{2}{@{}l}{\textbf{Loss}} \\
Energy-score MC samples & $16$ \\
Kinetic-action MC samples & $8$ \\
Kinetic regularisation & $10^{-8}$ \\

\bottomrule
\end{tabular}
\end{minipage}
\end{table}

\subsection{Ginzburg--Landau experiment}
\label{sec:appendix_experiments_gl}
\paragraph{Data generation.} The Ginzburg–Landau (GL) experiment is formulated as a parametric neural-field surrogate problem on the spatial domain $\Omega=[0,1]^2$, where the objective is to learn the mapping $(\mu,x,y)\mapsto u_{\mu}(x,y)$ from a scalar parameter $\mu\in[0,1]$ and spatial coordinates $(x,y)\in\Omega$ to the corresponding equilibrium field. Reference solutions are generated from the forced time-dependent GL equation
$$
\partial_t u=\varepsilon^2\Delta u+\mu u-u^3+f_{\mu}(x,y),
\qquad (x,y)\in\Omega,
$$
with homogeneous Neumann boundary conditions
$$
\nabla u\cdot n=0
\qquad\text{on }\partial\Omega,
$$
so that the equilibrium solution $u_\mu^\ast$ satisfies
$$
\varepsilon^2\Delta u_\mu^\ast+\mu u_\mu^\ast-(u_\mu^\ast)^3+f_\mu=0.
$$
The parameter-dependent forcing is constructed by continuously combining low-, intermediate- and high-frequency components,
$$
f_\mu=w_{\mathrm L}(\mu)f_{\mathrm L}
+w_{\mathrm M}(\mu)f_{\mathrm M}
+w_{\mathrm H}(\mu)f_{\mathrm H}
+0.12\cos\!\left[(2+3\mu)\pi x+2\pi\mu\right]
\cos\!\left[(5-2\mu)\pi y-2\pi\mu\right],
$$
where $w_{\mathrm L}=(1-2\mu)_+$, $w_{\mathrm H}=(2\mu-1)_+$, $w_{\mathrm M}=1-w_{\mathrm L}-w_{\mathrm H}$, and
$$
\begin{aligned}
f_{\mathrm L}&=0.40\cos(\pi x)+0.30\cos(2\pi y)
+0.20\cos(\pi x)\cos(\pi y),\\
f_{\mathrm M}&=0.35\cos(2\pi x+1.4\pi\mu)\cos(3\pi y-0.8\pi\mu)
+0.25\cos(3\pi x-2\pi\mu),\\
f_{\mathrm H}&=0.28\cos(4\pi x+2\pi\mu)\cos(2\pi y-\pi\mu)
-0.22\cos(5\pi y+1.6\pi\mu).
\end{aligned}
$$
The equation is discretised on a $64\times64$ Cartesian grid with $\varepsilon=0.04$ and integrated using a semi-implicit scheme with $\Delta t=0.2$,
$$
\left(I-\Delta t\,\varepsilon^2\Delta_h\right)u^{n+1}
=
u^n+\Delta t\left(\mu u^n-(u^n)^3+f_\mu\right).
$$
The initial state is $u^0=0.05f_\mu+0.15\sqrt{\mu}$, and iteration continues for at most $12\,000$ steps until
$$
\lVert u^{n+1}-u^n\rVert_\infty<10^{-6},
$$
with convergence checked every 25 steps; the final equilibrium is additionally assessed through
$$
r_{\mathrm{eq}}
=
\left\|
\varepsilon^2\Delta_hu^\ast+\mu u^\ast-(u^\ast)^3+f_\mu
\right\|_\infty.
$$

\begin{figure}
    \centering
    \includegraphics[width=0.8\linewidth]{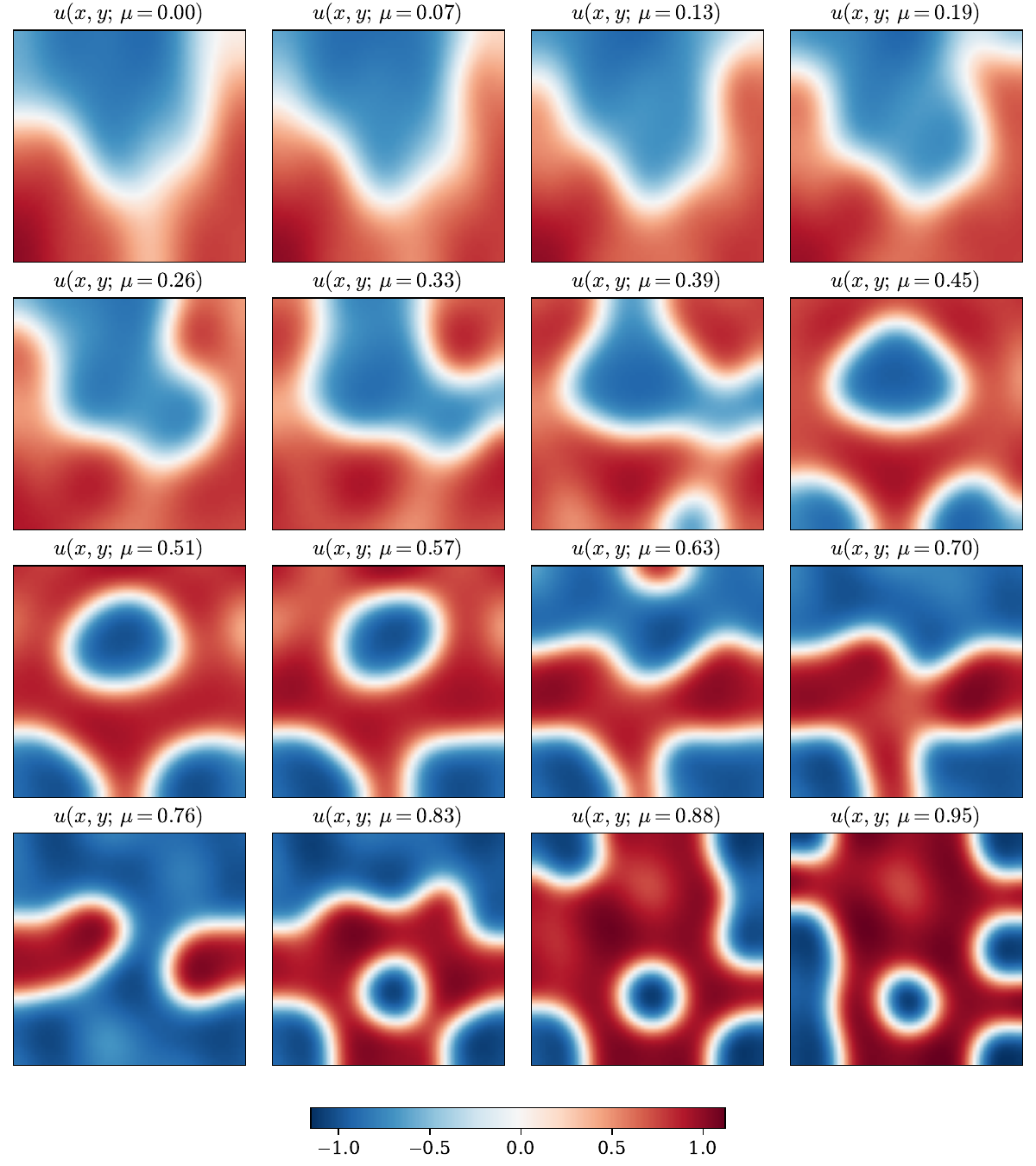}
    \caption{Ginzburg--Landau equilibrium fields experiment. Equilibrium field examples for 16 values of the input parameter $\mu$. As can be seen, the map $\mu \mapsto u_\mu$ is highly non-linear, albeit continuous.}
    \label{fig:gl-dataset}
\end{figure}

\paragraph{Dataset and hyperparameters.} The dataset contains 256 uniformly spaced values of $\mu$, split reproducibly into 204 training and 52 test fields. Each grid point is treated as an individual neural-field observation $((\mu,x,y),u_\mu^\ast(x,y))$. The deterministic baseline is a five-layer GeLU multilayer perceptron of width 128; vanilla dropout uses five layers of width $\lfloor128/(1-0.1)\rfloor=142$ with dropout probability $0.1$; and transported dropout uses width $128/0.5=256$, a Concrete mask distribution with $p=0.5$ and temperature $\tau=0.1$, and a two-layer velocity network of width 64 integrated using five Euler steps. Hidden widths are adjusted to match the expected number of active units under the reference masking distribution: approximately 128 units per layer for all three architectures. All models are trained for 2000 epochs with AdamW, learning rate $10^{-3}$, weight decay $10^{-5}$, and batch size $65\,536$, while predictive distributions are approximated using 32 stochastic forward passes. The setting is summarised in \autoref{tab:gl-hyperparameters}. The dataset is broadly represented by the fields in \autoref{fig:gl-dataset}.

\begin{table}
\centering
\footnotesize
\caption{Ginzburg--Landau equilibrium fields experiment. Experimental settings.}
\label{tab:gl-hyperparameters}
\begin{tabular}{ll}
\toprule
\textbf{Hyperparameter} & \textbf{Value} \\
\midrule
Hidden-layer depth & $5$ \\
Reference hidden width & $128$ \\
Deterministic hidden widths & $[128]\times5$ \\
Vanilla-dropout hidden widths & $[142]\times5$ \\
OT-dropout hidden widths & $[256]\times5$ \\
Activation & GeLU \\
Kernel initialization & He normal \\
Bias initialization & Zero \\
\midrule
Concrete probability $p$ & $0.5$ \\
Concrete temperature $\tau$ & $1.0$ \\
Velocity hidden layers & $[64,64]$ \\
Velocity activation & GeLU \\
Velocity initialization & He normal \\
ODE solver & Euler \\
Integration steps & $5$ \\
Energy Score samples $K_{\mathrm{ES}}$ & $4$ \\
Kinetic-action samples $K_{\mathrm{kin}}$ & $2$ \\
Kinetic weight $\lambda_{\mathrm{kin}}$ & $10^{-5}$ \\
\midrule
Dropout rate & $0.1$ \\
Deterministic/dropout loss & Mean squared error \\
OT-dropout loss & Energy Score $+$ kinetic action \\
Optimizer & AdamW \\
Learning rate & $10^{-3}$ \\
Weight decay & $10^{-5}$ \\
Training epochs & $2\,000$ \\
\toprule
\textbf{Other experimental settings} & \textbf{Value} \\
\midrule
Spatial domain & $\Omega=[0,1]^2$ \\
Spatial resolution & $64\times64$ \\
Input & $(\mu,x,y)\in\mathbb{R}^3$ \\
Output & $u_\mu(x,y)\in\mathbb{R}$ \\
Training cases & $204$ \\
Test cases & $52$ \\
Training fractions & $\{0.25,0.50,0.75,1.00\}$ \\
\midrule
Coverage levels & $\{0.50,0.75,0.80,0.90,0.95\}$ \\
Evaluation metrics & $L^2$, $L^1$, $S_{\text{ES}}$, MACE, PICP, Sharpness \\
\bottomrule
\end{tabular}
\end{table}

\paragraph{Results.} \autoref{tab:gl-per-sample} shows substantial heterogeneity across the test cases. Let $\epsilon_F$ be the first term of the energy score, and $\epsilon_D$ the second one, so that $S_{\text{ES}} = \epsilon_F - \frac 12 \epsilon_D$. Samples with small field error $\epsilon_F$ generally exhibit small Energy Score and predictive dispersion, whereas more difficult cases produce larger values of all three quantities. In particular, $\epsilon_D$ and the predictive standard deviation broadly track $\epsilon_F$ and the mean error, respectively, suggesting that predictive dispersion is informative about the spatial prediction error. The principal exception is sample $00255$, for which $\epsilon_F$ and the mean error increase markedly while $\epsilon_D$ and the standard deviation remain comparatively small. Thus, although uncertainty is generally aligned with error, the model appears under-dispersed on this particularly difficult case.

\begin{table}[ht]
\centering
\footnotesize
\caption{Per-sample results for the Ginzburg--Landau experiment.}
\label{tab:gl-per-sample}
\small
\begin{tabular}{l|ccccc}
\toprule
Sample & $\epsilon_F$ & $\epsilon_D$ & $S_{\text{ES}}$
& Mean error & Std. dev. \\
\midrule
00000 & 0.0242 & 0.0253 & 0.0116 & 0.0167 & 0.0268 \\
00011 & 0.0196 & 0.0232 & 0.0080 & 0.0114 & 0.0243 \\
00036 & 0.0192 & 0.0229 & 0.0077 & 0.0106 & 0.0228 \\
00050 & 0.0257 & 0.0277 & 0.0118 & 0.0167 & 0.0274 \\
00058 & 0.0208 & 0.0253 & 0.0082 & 0.0109 & 0.0265 \\
00077 & 0.0257 & 0.0288 & 0.0113 & 0.0171 & 0.0286 \\
00098 & 0.0597 & 0.0561 & 0.0317 & 0.0490 & 0.0550 \\
00116 & 0.0295 & 0.0351 & 0.0120 & 0.0164 & 0.0339 \\
00137 & 0.0289 & 0.0329 & 0.0124 & 0.0179 & 0.0311 \\
00160 & 0.0464 & 0.0431 & 0.0249 & 0.0360 & 0.0398 \\
00169 & 0.0339 & 0.0362 & 0.0158 & 0.0238 & 0.0335 \\
00187 & 0.0564 & 0.0622 & 0.0253 & 0.0403 & 0.0664 \\
00205 & 0.1094 & 0.1005 & 0.0591 & 0.0874 & 0.0927 \\
00229 & 0.1368 & 0.1160 & 0.0788 & 0.1163 & 0.1093 \\
00244 & 0.1207 & 0.0962 & 0.0727 & 0.0955 & 0.0895 \\
00255 & 0.1876 & 0.0925 & 0.1414 & 0.1647 & 0.0874 \\
\bottomrule
\end{tabular}
\end{table}

Interestingly, sample-level rank correlations provide information that is not captured by marginal calibration alone. As shown in \autoref{tab:gl-spearman} and \autoref{tab:gl-per-sample}, the field error $\epsilon_F$ is strongly correlated with the corresponding dispersion measure $\epsilon_D$. Hence, samples on which the surrogate incurs larger errors also tend to receive greater predictive dispersion. This indicates that the learned uncertainty is not merely calibrated in aggregate, but is also informative about the relative difficulty of individual test cases. Likewise, the strong association between the mean pointwise error and the predictive standard deviation suggests that this error--dispersion alignment persists at the local level. Nevertheless, such rank correlations measure the ordering of cases rather than the quantitative agreement between error and uncertainty; they should therefore be interpreted as complementary to, rather than implied by, calibration.

\begin{figure}
    \centering
    \includegraphics[width=0.8\linewidth]{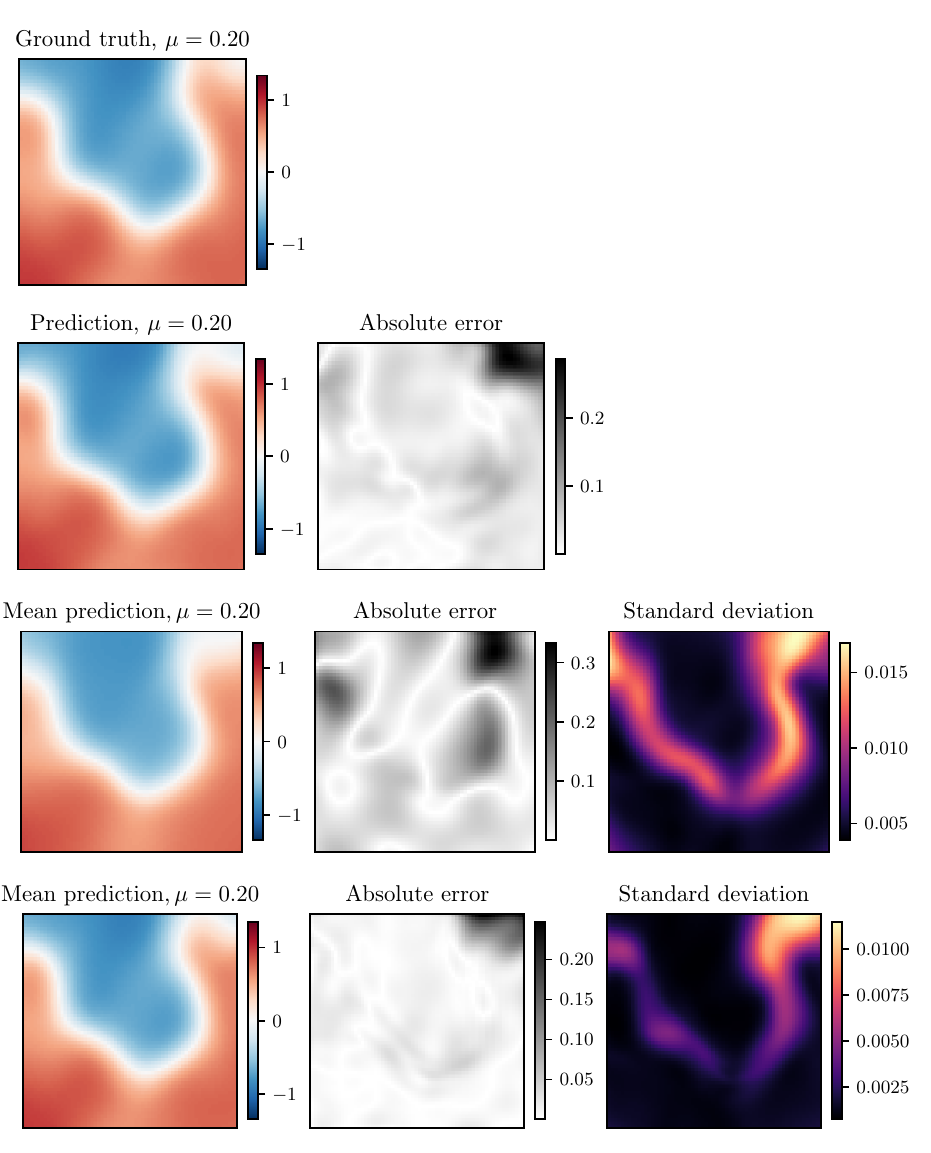}
    \caption{Ginzburg–Landau equilibrium fields experiment, sample 3. From top to bottom: ground truth, deterministic predictor, MC Dropout, OTD.}
    \label{fig:pred1}
\end{figure}

\begin{figure}
    \centering
    \includegraphics[width=0.8\linewidth]{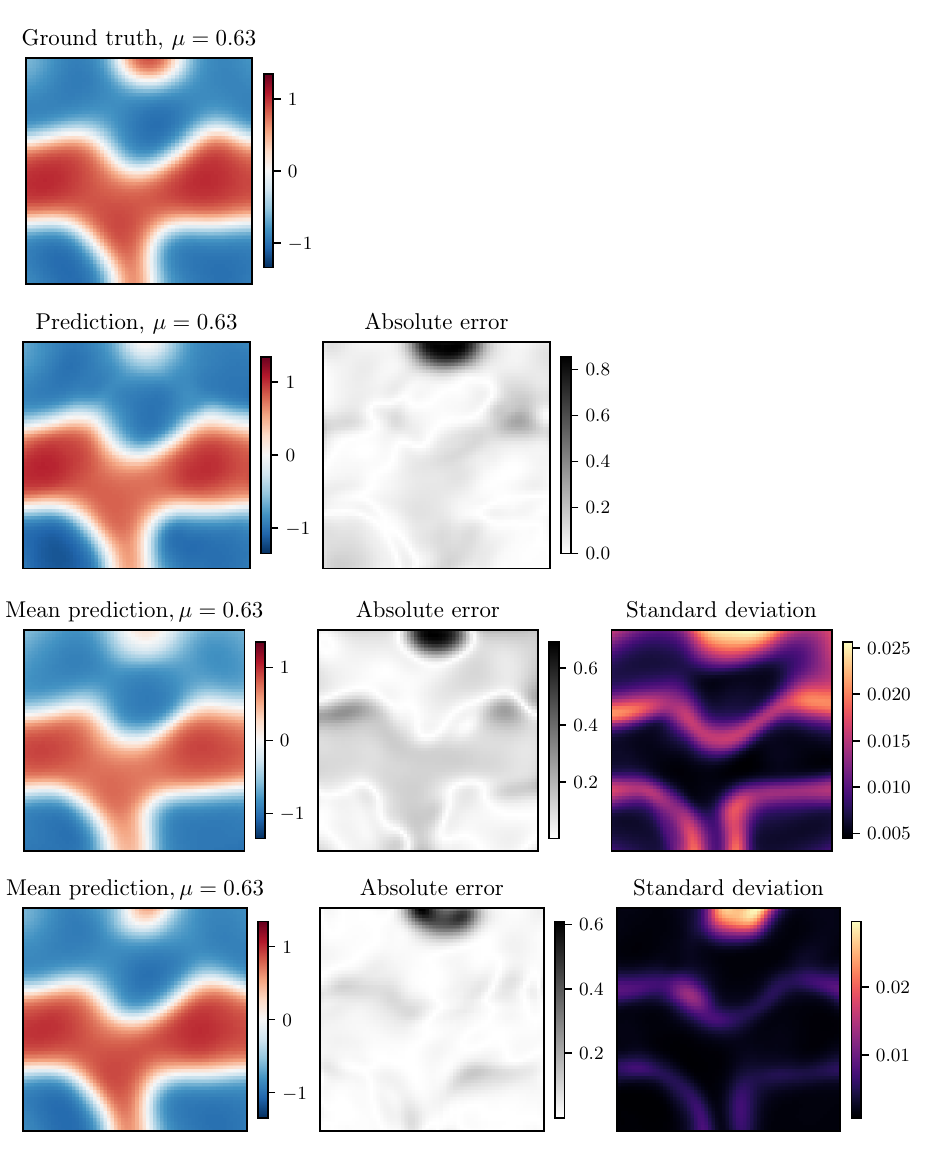}
    \caption{Ginzburg–Landau equilibrium fields experiment, sample 6. From top to bottom: ground truth, deterministic predictor, MC Dropout, OTD.}
    \label{fig:pred2}
\end{figure}

\begin{figure}
    \centering
    \includegraphics[width=0.8\linewidth]{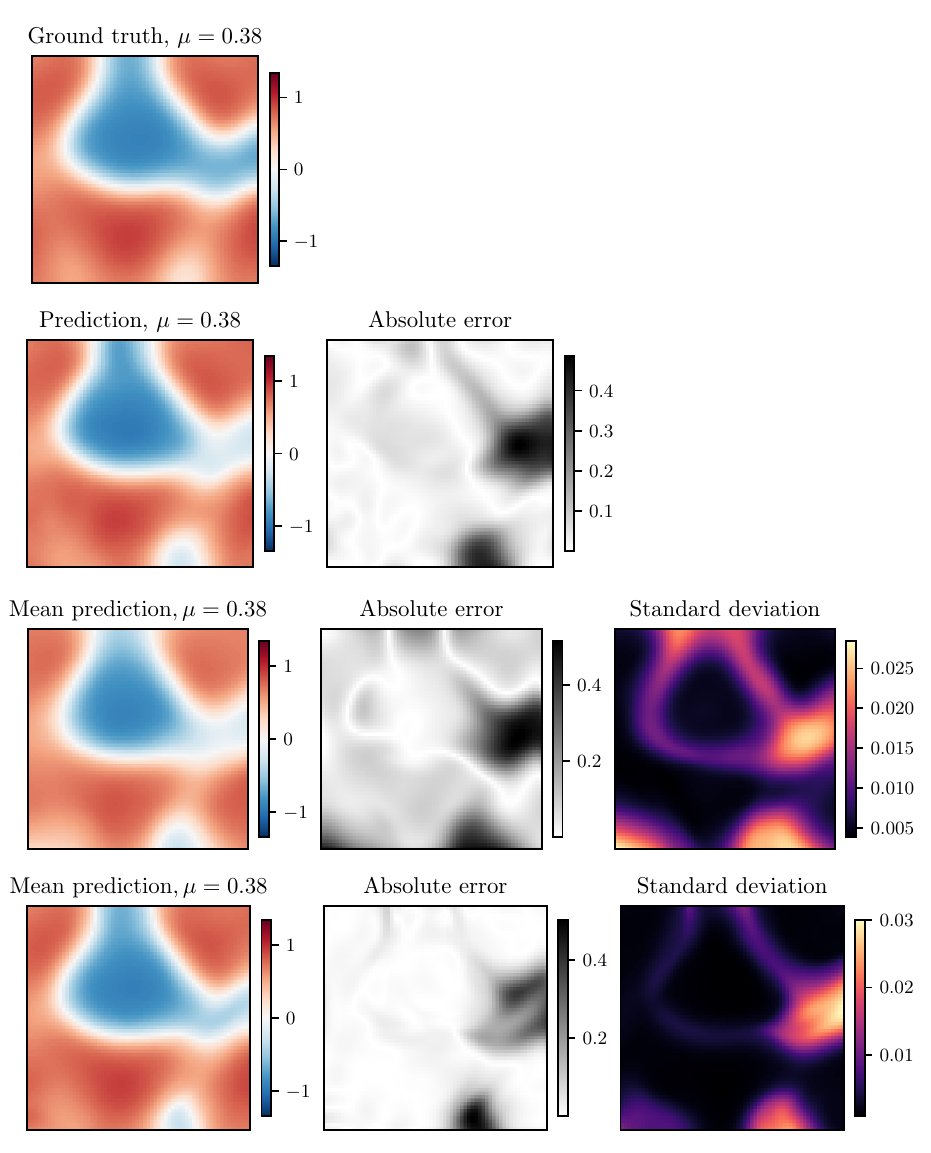}
    \caption{Ginzburg–Landau equilibrium fields experiment, sample 9. From top to bottom: ground truth, deterministic predictor, MC Dropout, OTD.}
    \label{fig:pred3}
\end{figure}

\begin{figure}
    \centering
    \includegraphics[width=0.8\linewidth]{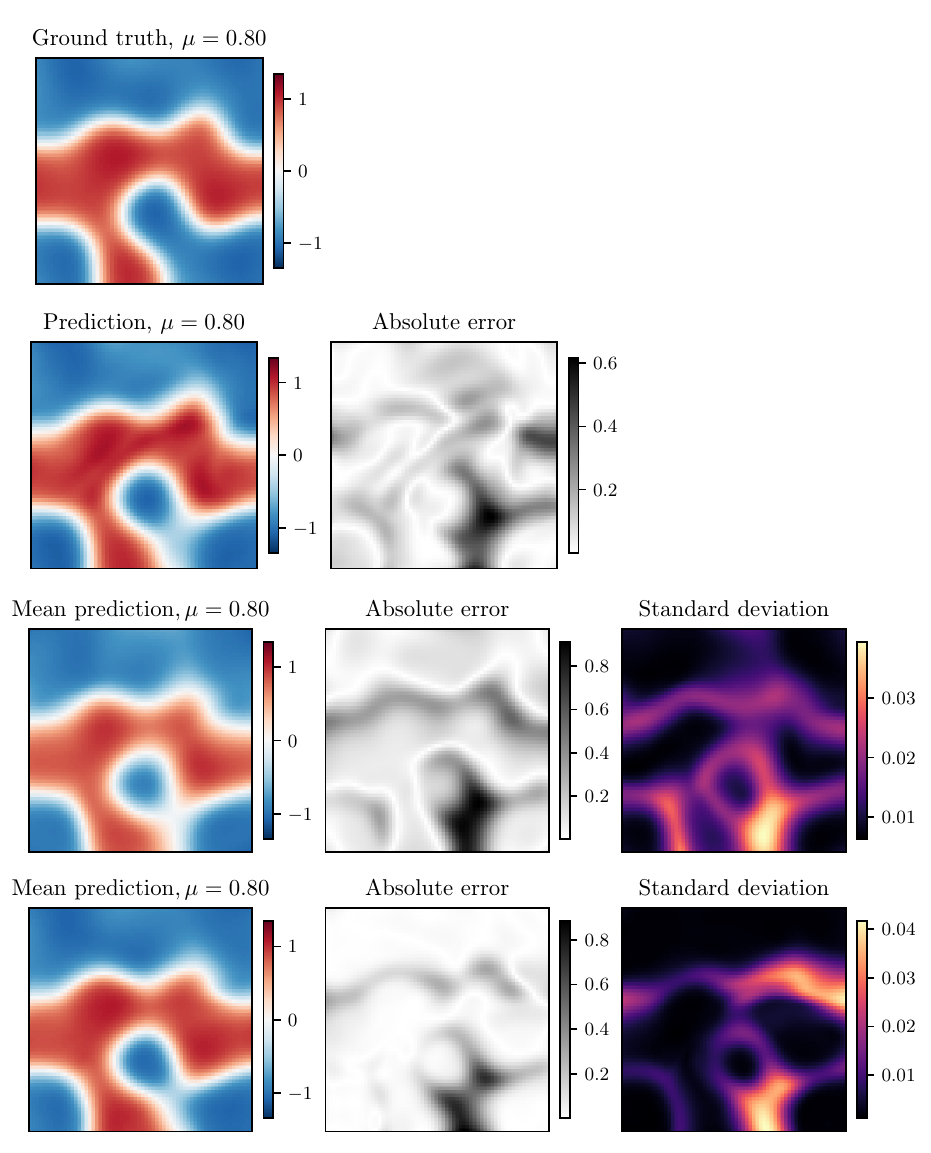}
    \caption{Ginzburg–Landau equilibrium fields experiment, sample 12. From top to bottom: ground truth, deterministic predictor, MC Dropout, OTD.}
    \label{fig:pred4}
\end{figure}

Some sample predictions are shown in \autoref{fig:pred1}--\autoref{fig:pred4}.

\begin{figure}
    \centering
    \includegraphics[width=0.8\linewidth]{img/energy_maps_06_gl_00098.pdf}
    \includegraphics[width=0.8\linewidth]{img/energy_maps_08_gl_00137.pdf}
    \includegraphics[width=0.8\linewidth]{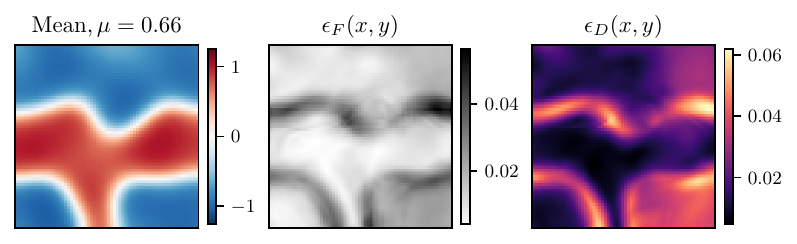}
    \includegraphics[width=0.8\linewidth]{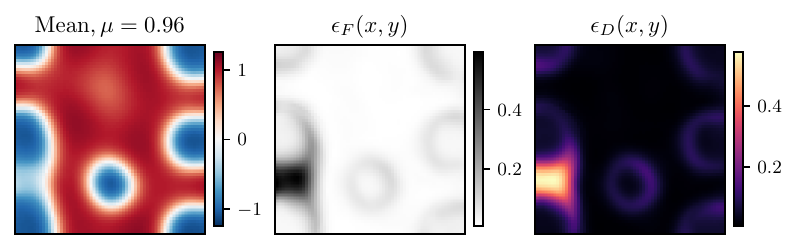}
    \caption{Ginzburg–Landau equilibrium fields experiment. Spatial alignment between the first ($\epsilon_F$) and the second term ($\epsilon_D$) of the Energy score.}
    \label{fig:align}
\end{figure}

\autoref{fig:align} shows the spatial alignment between the first and the second term of the Energy Score.

\begin{table}[ht]
\centering
\footnotesize
\caption{Ginzburg–Landau equilibrium fields experiment. Pairwise Spearman rank correlations.}
\label{tab:gl-spearman}
\small
\begin{tabular}{l|cccc}
\toprule
& $\varepsilon_F$
& $\varepsilon_D$
& Mean error
& Std. dev. \\
\midrule
$\varepsilon_F$ & 1      & \textbf{0.9765} & --     & 0.9735 \\
$\varepsilon_D$ & \textbf{0.9765} & 1      & 0.9441 & --     \\
Mean error      & --     & 0.9441 & 1      & 0.9294 \\
Std. dev.       & 0.9735 & --     & 0.9294 & 1      \\
\bottomrule
\end{tabular}
\end{table}

This relationship becomes clearer at the pointwise level, when $\epsilon_F(x,y)$ and $\epsilon_D(x,y)$ are treated as fields over the unit square rather than aggregated over the domain. In particular, \autoref{tab:gl-spearman-pixel} and \autoref{tab:gl-spearman-sample} show that, when the model is well calibrated, $\epsilon_F$ is strongly correlated with $\epsilon_D$. This association is consistent with the Energy Score objective, which jointly rewards agreement with the observation and appropriate ensemble dispersion. Indeed, if the optimal conditional law coincides with the target one, the two terms coincide. Consequently, if each realisation $y^{(k)}$ is assigned a physical interpretation, the mean realisation-wise error $\varepsilon_F$ is more representative than the error of the ensemble mean $e$, as the latter may conceal compensating errors across realisations. For the same reason, the pairwise dispersion $\epsilon_D$ provides a more informative proxy for the reliability of $\varepsilon_F$ than the pointwise standard deviation $\sigma$.

\begin{table}[ht]
\centering
\footnotesize
\caption{Ginzburg–Landau equilibrium fields experiment. Per-case pixel-wise Spearman rank correlations. 
The largest correlation in each row is shown in bold.}
\label{tab:gl-spearman-pixel}
\begin{tabular}{l|cccc}
\toprule
Sample
& $\rho(\varepsilon_F,\varepsilon_D)$
& $\rho(e,\sigma)$
& $\rho(\varepsilon_F,\sigma)$
& $\rho(e,\varepsilon_D)$ \\
\midrule
00000 & \textbf{0.6290} &  0.1138 & 0.5881 &  0.1264 \\
00011 & \textbf{0.6909} & -0.0576 & 0.6559 & -0.0404 \\
00036 & \textbf{0.7722} &  0.1702 & 0.7455 &  0.1886 \\
00050 & \textbf{0.8393} &  0.3215 & 0.8049 &  0.3426 \\
00058 & \textbf{0.8659} &  0.3631 & 0.8326 &  0.3810 \\
00077 & \textbf{0.9168} &  0.4516 & 0.8773 &  0.4826 \\
00098 & \textbf{0.9439} &  0.7192 & 0.9284 &  0.7335 \\
00116 & \textbf{0.9083} &  0.4198 & 0.8886 &  0.4286 \\
00137 & \textbf{0.9050} &  0.4510 & 0.8962 &  0.4659 \\
00160 & \textbf{0.9314} &  0.6722 & 0.9289 &  0.6747 \\
00169 & \textbf{0.8957} &  0.5480 & 0.8888 &  0.5558 \\
00187 & \textbf{0.9229} &  0.6190 & 0.8966 &  0.6525 \\
00205 & \textbf{0.9686} &  0.8180 & 0.9675 &  0.8180 \\
00229 & \textbf{0.9688} &  0.8185 & 0.9631 &  0.8225 \\
00244 & \textbf{0.9622} &  0.7183 & 0.9562 &  0.7255 \\
00255 & \textbf{0.5934} &  0.2675 & 0.5639 &  0.2925 \\
\bottomrule
\end{tabular}
\end{table}

\begin{table}[ht]
\centering
\footnotesize
\caption{Pairwise Spearman rank correlations across samples.}
\label{tab:gl-spearman-sample}
\small
\begin{tabular}{l|cccc}
\toprule
& $\varepsilon_F$
& $\varepsilon_D$
& Mean error
& Std. dev. \\
\midrule
$\varepsilon_F$ & 1      & \textbf{0.9263} & --     & 0.9095 \\
$\varepsilon_D$ & \textbf{0.9263} & 1      & 0.6085 & --     \\
Mean error      & --     & 0.6085 & 1      & 0.5926 \\
Std. dev.       & 0.9095 & --     & 0.5926 & 1      \\
\bottomrule
\end{tabular}
\end{table}

\paragraph{Disentangling the contributions of the Energy Score and transport flow.} The comparison is conducted using $50\%$ of the available training set, corresponding to 102 equilibrium fields, and a fixed $50\%$ subset of the test set, corresponding to 26 fields. The same test subset is used throughout, while the training subset, model initialisation, data ordering, and stochastic keys are paired across variants within each random seed. We considered five configurations: MC Dropout trained with mean squared error; MC Dropout trained with the Energy Score; zero-step OTD trained with the Energy Score; OTD with an active flow but $\lambda_{\mathrm{kin}}=0$; and complete OTD with kinetic regularisation. All predictive networks have four GeLU hidden layers and are scaled to retain approximately 128 active units per layer on average: MC Dropout uses width $\lfloor128/(1-0.1)\rfloor=142$ with dropout probability $0.1$, whereas the OTD variants use width $128/0.5=256$, a Concrete reference distribution with $p=0.5$, and temperature $\tau=1.0$. The zero-step variant applies no transport, while the flow-based variants use a two-layer velocity network of width 64 and two Euler integration steps. The complete model uses kinetic weight $\lambda_{\mathrm{kin}}=10^{-7}$, with $K_{\mathrm{ES}}=4$ predictive samples and $K_{\mathrm{kin}}=2$ kinetic-action samples during training. We train every configuration for 800 epochs with AdamW, learning rate $10^{-3}$, weight decay $10^{-5}$, and batches of 16 fields, corresponding to $65\,536$ spatial observations. Evaluation uses 128 stochastic forward passes and reports RMSE and MAE errors, Energy Score, and MACE. Results are aggregated over three random seeds.

\paragraph{Sensitivity to the integration depth and Concrete temperature.}
We examine the sensitivity of OTD to the number of flow-integration steps $L$ and the temperature of the reference Concrete distribution $\tau$. As shown in \autoref{tab:integration_steps_sensitivity}, the largest improvement occurs between zero and one step: MAE decreases from $0.0949$ to $0.0555$, while MACE decreases from $0.1736$ to $0.0367$. Increasing the integration depth produces smaller but consistent gains, with four steps achieving the best result across all predictive metrics.

\begin{table}[h]
    \centering
    \caption{Sensitivity to the number of integration steps at $\tau=1$. Results are mean $\pm$ standard deviation over three seeds.}
    \label{tab:integration_steps_sensitivity}
    \footnotesize
    \begin{tabular}{c|cccc}
        \toprule
        $L$
        & RMSE
        & MAE
        & $S_{\mathrm{ES}}$
        & MACE \\
        \midrule
        $0$
            & $0.1704 \pm 0.0138$
            & $0.0949 \pm 0.0074$
            & $0.0698 \pm 0.0027$
            & $0.1736 \pm 0.0132$ \\
        $1$
            & $0.1417 \pm 0.0188$
            & $0.0555 \pm 0.0086$
            & $0.0399 \pm 0.0071$
            & $0.0367 \pm 0.0309$ \\
        $2$
            & $0.1392 \pm 0.0199$
            & $0.0542 \pm 0.0065$
            & $0.0386 \pm 0.0053$
            & $0.0255 \pm 0.0095$ \\
        $4$
            & $\mathbf{0.1368 \pm 0.0098}$
            & $\mathbf{0.0533 \pm 0.0043}$
            & $\mathbf{0.0377 \pm 0.0028}$
            & $\mathbf{0.0105 \pm 0.0072}$ \\
        \bottomrule
    \end{tabular}
\end{table}

The temperature results in \autoref{tab:temperature_sensitivity} show that references closer to the Bernoulli limit produce larger calibration errors and activate more units on average. Increasing the temperature improves the probabilistic metrics: $\tau=1$ achieves the lowest MAE, Energy Score, and MACE, while $\tau=0.5$ achieves the lowest RMSE, albeit with greater variability across seeds. Overall, the results favour a smoother reference distribution and show that OTD is relatively stable for $\tau\in\{0.5,1\}$.

\begin{table}[h]
    \centering
    \caption{Sensitivity to the Concrete temperature using two integration steps. Results are mean $\pm$ standard deviation over three seeds.}
    \label{tab:temperature_sensitivity}
    \footnotesize
    \begin{tabular}{c|cccc}
        \toprule
        $\tau$
        & RMSE
        & MAE
        & $S_{\mathrm{ES}}$
        & MACE \\
        \midrule
        $0.05$
            & $0.1412 \pm 0.0133$
            & $0.0588 \pm 0.0044$
            & $0.0422 \pm 0.0029$
            & $0.0458 \pm 0.0196$ \\
        $0.10$
            & $0.1400 \pm 0.0048$
            & $0.0601 \pm 0.0013$
            & $0.0422 \pm 0.0014$
            & $0.0453 \pm 0.0447$ \\
        $0.50$
            & $\mathbf{0.1372 \pm 0.0245}$
            & $0.0546 \pm 0.0084$
            & $0.0387 \pm 0.0068$
            & $0.0344 \pm 0.0146$ \\
        $1.00$
            & $0.1392 \pm 0.0199$
            & $\mathbf{0.0542 \pm 0.0065}$
            & $\mathbf{0.0386 \pm 0.0053}$
            & $\mathbf{0.0255 \pm 0.0095}$ \\
        \bottomrule
    \end{tabular}
\end{table}

\subsection{UCI/OpenML Protocol}
\label{sec:appendix_experiments_openml}
We evaluate the method on a collection of UCI/OpenML\footnote{Datasets were retrieved from OpenML: Energy Efficiency (ID 43918), Synchronous Machine (44968), CT Slice Localisation (42973), QSAR Fish Toxicity (44028), Concrete Compressive Strength (4353), Concrete Slump (41490), Forest Fires (46587), Naval Propulsion (44969), Infrared Thermography (46613), SUPPORT2 (46860), and Superconductivity (43174). Source: OpenML; Vanschoren et al., “OpenML: Networked Science in Machine Learning”, SIGKDD Explorations, 2013.} tabular regression datasets, including Energy, Concrete, Slump, Forest Fires, Naval Propulsion, Toxicity, Synchronous Machine, Localisation, Thermography, and Superconductivity. These datasets are based on real experimental or observational measurements, and therefore include measurement noise, finite-sample effects, and possible heteroscedasticity. This makes them a useful test bed for predictive uncertainty, since a good model should not only fit the conditional mean but also provide calibrated predictive distributions.

For each dataset, we use a five-fold cross-validation protocol. Within each training fold, $10\%$ of the data is held out as a validation set for early stopping and for selecting the density-estimation hyperparameters; test labels are not used for tuning. Inputs and targets are standardised fold-wise, with a log transform applied to \texttt{forestfires} and \texttt{supercond} datasets. The Optimal Transport Dropout model produces $K=1000$ predictive samples per test point, from which we estimate likelihood-based metrics, RMSE, $S_{\text{ES}}$, prediction interval coverage, sharpness, and calibration errors. Overall, the results are favourable: the method provides competitive point prediction accuracy while producing meaningful predictive distributions, and several cases show good calibration across noisy real-world regression tasks.

This is a relevant experiment since it shows the capability of the model beyond high-fidelity data. In particular, it shows that it is also able to model noisy data, capturing aleatoric uncertainty stemming from experimental measurement noise and unobserved covariates. Because OTD defines an implicit predictive distribution rather than an analytic density, we estimate its NLL from predictive samples using a Gaussian kernel density estimator. We therefore regard NLL as a complementary, density-estimator-dependent metric and report sample-based proper scores and calibration diagnostics alongside it (see \autoref{tab:uci}).

\begin{table}[ht]
    \centering
    \caption{Experimental dataset. Results reported as mean NLL. The better, the darker.}
    \label{tab:results}
    \footnotesize
    \setlength{\tabcolsep}{4pt}
    \resizebox{\linewidth}{!}{%
        \begin{tabular}{lcccccccccccc}
            \toprule
            Dataset & CADET & CART-h & CART-k & CKDE & LSCDE & NKDE & CDE-HT & LinCDE & MDN & NF & CPFN & OTD\\
            \midrule
            energy      & 3.55  & 3.09 & 3.06 & \cellcolor{bluemodern!20}2.47 & 3.38 & 3.00 & 2.93 & 2.93 & 2.78 & 2.86 & \cellcolor{bluemodern!10}2.73 & \cellcolor{bluemodern!30}1.08\\
            
            synchronous  & -2.93 & -1.63 & -1.86 & \cellcolor{bluemodern!10}-3.59 & -1.25 & -1.57 & -2.11 & -1.85 & -2.94 & -2.64 & \cellcolor{bluemodern!20}-4.48 & \cellcolor{bluemodern!30}-5.02\\
            
            localization  & -0.23 & -0.55 & -0.01 & -0.26 & -0.61 & -0.28 & \cellcolor{bluemodern!10}-0.66 & \cellcolor{bluemodern!30}-0.95 & \cellcolor{bluemodern!20}-0.68 & -0.43 & 4.19 & 1.03\\
            
            toxicity    & 1.80  & 1.50 & 1.38 & 1.32 & 1.34 & 1.55 & 1.53 & \cellcolor{bluemodern!10}1.29 & \cellcolor{bluemodern!20}1.24 & \cellcolor{bluemodern!30}1.23 & 2.13 & 1.40\\
            
            concrete    & 4.17  & 3.75 & 3.93 & 3.32 & 3.66 & 3.91 & 3.72 & 3.47 & \cellcolor{bluemodern!30}2.97 & \cellcolor{bluemodern!10}3.18 & 3.40 & \cellcolor{bluemodern!20}3.07\\
            
            slump       & 3.42  & 3.55 & 3.43 & \cellcolor{bluemodern!20}2.35 & 2.91 & 3.08 & 3.34 & 2.98 & \cellcolor{bluemodern!30}2.23 & \cellcolor{bluemodern!10}2.39 & 4.48 & 2.40\\
            
            forestfire  & 133.95 & 3.96 & 4.39 & 4.85 & 4.68 & 5.55 & 3.43 & 4.35 & \cellcolor{bluemodern!10}3.26 & \cellcolor{bluemodern!20}3.23 & \cellcolor{bluemodern!30}2.53 & 3.50\\
            
            navalpropulsion  & -3.53 & -3.30 & -3.66 & -2.8 & -2.88 & -3.19 & -3.60 & -3.36 & \cellcolor{bluemodern!10}-4.12 & -3.75 & \cellcolor{bluemodern!30}-5.61 & \cellcolor{bluemodern!20}-5.38\\
            
            thermograpy  & 2.21  & 0.66 & 0.72 & 0.66 & 0.94 & 0.94 & 0.64 & 0.59 & \cellcolor{bluemodern!10}0.56 & \cellcolor{bluemodern!20}0.52 & 0.91 & \cellcolor{bluemodern!30}0.18\\

            support2    & 97.3 & 0.51 & \cellcolor{bluemodern!10} 0.32 & 2.09 & 2.46 & 2.13 & \cellcolor{bluemodern!20}0.29 & 1.48 & 1.53 & 1.24 &  1.57 & \cellcolor{bluemodern!30} -0.19 \\
            
            superconductor  & 9.6   & 3.84 & 4.36 & 4.55 & 4.17 & 4.19 & \cellcolor{bluemodern!10}3.48 & 3.87 & \cellcolor{bluemodern!30}3.33 & 3.50 & 3.51 & \cellcolor{bluemodern!20}3.45\\
            \bottomrule
        \end{tabular}
    }
\end{table}

\begin{table}[ht]
\caption{Experimental dataset. Summary of the OTD results across datasets. MACE is computed against nominal coverage levels $\alpha \in \{0.50, 0.80, 0.90, 0.95\}$.}
\label{tab:uci}
\centering
\footnotesize
\setlength{\tabcolsep}{5pt}
\begin{tabular}{lccccc}
\toprule
Dataset & RMSE  & $S_{\text{ES}}$ & Sharpness & MACE \\
\midrule
energy          & 0.857 & 0.464 & 1.231 & 0.323 \\
synchronous     & 0.001 & 0.001 & 0.006 & 0.064 \\
localization    & 0.829 & 0.395 & 1.866 & 0.142 \\
toxicity        & 1.028 & 0.569 & 1.605 & 0.278 \\
concrete        & 5.172 & 2.965 & 6.830 & 0.355 \\
slump           & 2.345 & 1.613 & 0.721 & 0.690 \\
forestfires     & 1.924 & 1.397 & 0.416 & 0.717 \\
navalpropulsion & 0.002 & 0.001 & 0.006 & 0.066 \\
thermography    & 1.242 & 0.262 & 0.153 & 0.640 \\
support2        & 0.296 & 0.121 & 0.489 & 0.046 \\
supercond       & 0.426 & 0.206 & 1.165 & 0.067 \\
\bottomrule
\end{tabular}
\end{table}

\section{The role of the kinetic regularisation}
\label{sec:appendix_kinetic}
\begin{figure}[ht]
    \centering
    \includegraphics[width=0.7\linewidth]{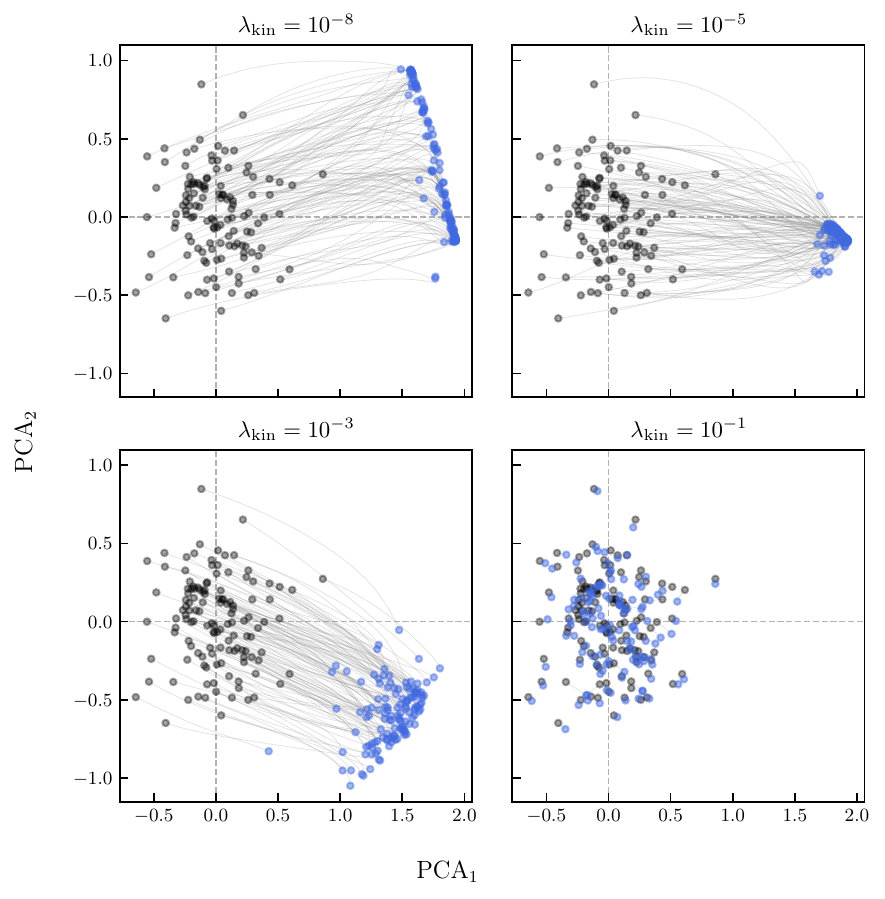}
    \caption{Role of the kinetic regularisation. Grey dots denote the reference masks, while blue dots denote the transported masks. Larger values of $\lambda_{\mathrm{kin}}$ produce smoother transport trajectories but increasingly restrict the deformation of the reference measure.}
    \label{fig:kinetic_regularisation}
\end{figure}

The kinetic regularisation weight $\lambda_{\mathrm{kin}}$ directly controls the transport magnitude of the latent trajectories. As $\lambda_{\mathrm{kin}}$ increases, the learned velocity field is encouraged to minimise its kinetic action, yielding progressively smoother paths in latent space. However, excessively strong regularisation also penalises the transport itself, limiting the deformation of the reference measure and thus reducing the expressiveness of the resulting push-forward distribution. Consequently, the model may struggle to represent complex target distributions, leading to underfitting. Therefore, $\lambda_{\mathrm{kin}}$ must be chosen as a compromise between latent trajectory regularity and transport flexibility. Figure~\ref{fig:kinetic_regularisation} illustrates this trade-off on the bimodal experiment: increasing $\lambda_{\mathrm{kin}}$ produces progressively smoother trajectories, whereas overly large values inhibit the transport and reduce the quality of the learned distribution. In practice, we choose $\lambda_{\mathrm{kin}}$ so that the $\lambda_{\mathrm{kin}} \mathcal A(\psi)$ remains small relative to the Energy Score during training, and check that increasing it does not materially worsen the validation Energy Score. This limits the cost of transport without unduly compromising predictive fit.

\section{Sensitivity analysis: timing sweep}
\label{sec:appendix_timing}
We performed an OTD-only timing ablation on the Ginzburg--Landau experiment to identify which transported-dropout components contribute most to wall-clock training time. All runs used the same training subset, namely half of the available GL training cases, batch size $8192$, and $1000$ measured optimisation epochs. The reference configuration used $K_{\mathrm{ES}}=4$ Monte Carlo samples for the energy score, $K_{\mathrm{kin}}=2$ samples for the kinetic regularizer, $2$ Euler integration steps in the transport flow, and the full $64\times64$ spatial grid. We then varied one factor at a time: $K_{\mathrm{ES}}\in\{1,2,4,8\}$, $K_{\mathrm{kin}}\in\{0,1,2,4\}$, the number of integration steps in $\{0,1,2,4\}$, and the grid resolution in $\{64,32,16\}$. The case $K_{\mathrm{ES}}=1$ is included as a limiting computational baseline: with a single predictive sample, the pairwise-dispersion term of the empirical energy score vanishes, leaving only the sample-to-observation discrepancy. Timings exclude the first $5$ warmup epochs for each model, to avoid measuring JAX/XLA compilation overhead, and are averaged over three initialisation seeds.

\begin{table}[t]
\centering
\caption{OTD timing-sweep configurations for the Ginzburg--Landau
experiment. All variants were trained using $50\%$ of the training cases
and two cases per batch. Timings are reported as mean $\pm$ standard
deviation over three seeds. Each run used $5$ warmup epochs, excluded
from timing, followed by $1000$ measured training epochs. Speedups are
computed relative to the reference configuration.}
\label{tab:timesweep}
\footnotesize
\begin{adjustbox}{max width=\linewidth}
\begin{tabular}{lcccccc}
\toprule
Configuration
& $K_{\mathrm{ES}}$
& $K_{\mathrm{kin}}$
& Integration steps
& Grid
& Time (s)
& Speedup \\
\midrule
Reference
& 4 & 2 & 2 & $64^2$ & $145.76 \pm 2.36$ & $1.00\times$ \\
\midrule
$K_{\mathrm{ES}}=1$
& 1 & 2 & 2 & $64^2$ & $149.22 \pm 1.53$ & $0.98\times$ \\
$K_{\mathrm{ES}}=2$
& 2 & 2 & 2 & $64^2$ & $149.22 \pm 4.52$ & $0.98\times$ \\
$K_{\mathrm{ES}}=8$
& 8 & 2 & 2 & $64^2$ & $150.35 \pm 0.82$ & $0.97\times$ \\
\midrule
No kinetic term
& 4 & 0 & 2 & $64^2$ & $140.87 \pm 5.01$ & $1.03\times$ \\
$K_{\mathrm{kin}}=1$
& 4 & 1 & 2 & $64^2$ & $146.21 \pm 7.13$ & $1.00\times$ \\
$K_{\mathrm{kin}}=4$
& 4 & 4 & 2 & $64^2$ & $150.59 \pm 2.79$ & $0.97\times$ \\
\midrule
Zero integration steps
& 4 & 2 & 0 & $64^2$ & $137.87 \pm 6.51$ & $1.06\times$ \\
One integration step
& 4 & 2 & 1 & $64^2$ & $138.78 \pm 1.42$ & $1.05\times$ \\
Four integration steps
& 4 & 2 & 4 & $64^2$ & $145.05 \pm 5.46$ & $1.00\times$ \\
\midrule
Grid $32\times32$
& 4 & 2 & 2 & $32^2$ & $148.82 \pm 3.28$ & $0.98\times$ \\
Grid $16\times16$
& 4 & 2 & 2 & $16^2$ & $144.81 \pm 1.14$ & $1.01\times$ \\
\bottomrule
\end{tabular}
\end{adjustbox}
\end{table}

The observed timings indicate that, at full grid resolution, runtime is only weakly sensitive to the Monte Carlo counts: reducing $K_{\mathrm{ES}}$ from $4$ to $1$ did not reduce wall-clock time, while increasing it to $8$ produced a comparable time. Predictive samples and spatial points are processed in batches and can be evaluated in parallel when sufficient hardware resources are available. Removing the kinetic term yielded a modest $1.03\times$ speedup. Flow evaluations, however, are sequential across integration steps; using zero or one step yielded speedups of $1.06\times$ and $1.05\times$, respectively, while four steps gave a time comparable to the reference. Thus, the measured contribution of flow integration to total runtime is modest in this configuration.

\begin{figure}
    \centering
    \includegraphics[width=0.9\linewidth]{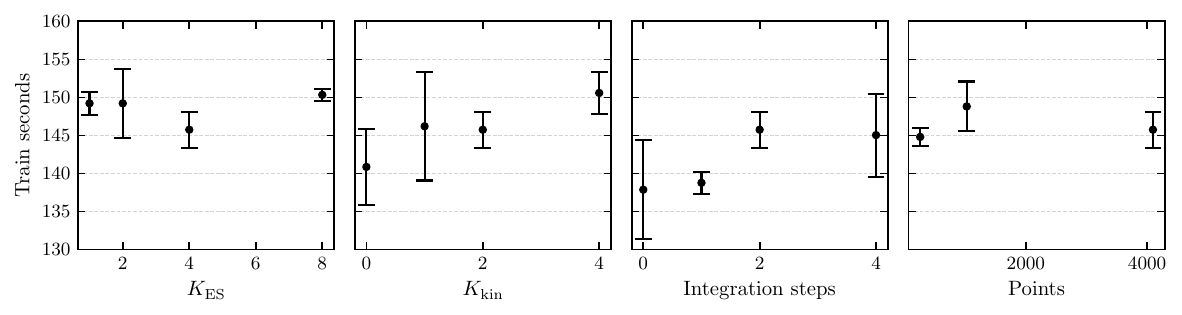}
    \caption{OTD timing-sweep configurations for the Ginzburg--Landau experiment. All variants were trained using $50\%$ of the training cases and batch size $8192$. Timings are reported as mean $\pm$ standard deviation over three seeds. Each run used $5$ warmup epochs, excluded from timing, followed by $1000$ measured training epochs.}
    \label{fig:timesweep}
\end{figure}

Reducing the spatial grid likewise produced no substantial speedup in the reported measurements: the $32\times32$ and $16\times16$ configurations took $148.82$ and $144.81$ seconds, respectively, compared with $145.76$ seconds for the $64\times64$ reference. This is consistent with parallel evaluation of batched grid points on the NVIDIA GeForce RTX 5090 used for all timing runs, although the timings alone do not establish which operation dominates computational cost. The small measured overhead of the flow permits the use of a few integration steps; kinetic regularisation discourages large latent velocities, but does not guarantee that the resulting time discretisation is accurate. Its adequacy should be assessed by comparing the learned predictions with those obtained using a finer integration grid. The results are reported in \autoref{tab:timesweep} and \autoref{fig:timesweep}.

\section{A generalisation of OTD}
\label{sec:appendix_generalisation}

In this section we discuss a generalisation of OTD in which the transported random variable is no longer tied to the width of the stochastic feature mask. In the original OTD construction, randomness enters through a transported dropout mask. If the predictive network has hidden widths
$(h_1,\ldots,h_L)$, the transported state has dimension
$$
d_z = \sum_{\ell=1}^L h_\ell .
$$
Thus the latent dimension is determined by the architecture of the downstream network. This coupling is natural for mask modulation, but it also makes the transport problem more expensive as the predictive model is widened or deepened.

\paragraph{Conditional formulation.} The conditional formulation decouples these two objects. We sample a latent variable $z_0 \sim \rho_0$ transport it through a learned flow $z_1 = T_\theta(z_0),$ and concatenate the transported latent code to the input of the neural field. Here $d_z$ is a free modelling choice. It is no longer required to coincide with the total number of hidden features in the network. This makes the construction strictly more general: the stochastic component can be made large enough to represent epistemic uncertainty, but small enough to keep the transport problem
tractable.

\paragraph{Low--latent formulation.} Instead, the low--latent formulation prescribes the latent dimension $d_z$ a priori. This keeps the latent flow minimal in the latent space of dimension $d_z$. To match the predictive network's dimensionality, we then pass the flow output to a neural network that lifts the latent mask to a higher-dimensional space, matching the predictive network's actual dimension, $\sum_{\ell = 1}^L h_\ell$.

\paragraph{Experimental results}
In the GL experiment, the original OTD mask dimension is $d_z = 6 \cdot 256 = 1536$. We therefore compare the matched latent dimension against lower-dimensional conditional models with $d_z \in \{512,128,32\}$. The model is trained with a single latent code per field realisation, broadcast over all spatial points of that case.

\begin{table}[t]
\centering
\caption{Ginzburg--Landau results comparing the original OTD experiment with
the low-dimensional OTD variants and Conditional OT variants.
RMSE, MAE, $S_{\text{ES}}$, MACE, sharpness, and $\mathrm{PICP}_{0.90}$ are evaluated
at the pixel level on a $64\times64$ grid. MACE is averaged over
$\alpha \in \{0.50,0.75,0.80,0.90,0.95\}$.}
\label{tab:gl-low-latent}
\footnotesize
\resizebox{\linewidth}{!}{%
\begin{tabular}{lrrrrrrr}
\toprule
Method
& $d_z$
& RMSE
& MAE
& $S_{\text{ES}}$
& MACE
& Sharpness
& $\mathrm{PICP}_{0.90}$ \\
\midrule
OTD
& 1536
& $\mathbf{0.134 \pm 0.003}$
& $\mathbf{0.043 \pm 0.000}$
& $\mathbf{0.031 \pm 0.000}$
& $\mathbf{0.038 \pm 0.020}$
& $0.046 \pm 0.005$
& $\mathbf{0.911 \pm 0.024}$ \\
\midrule
Low-latent OTD
& 512
& $0.140 \pm 0.003$
& $0.045 \pm 0.001$
& $0.033 \pm 0.001$
& $0.086 \pm 0.038$
& $0.039 \pm 0.004$
& $0.811 \pm 0.031$ \\
Low-latent OTD
& 128
& $0.141 \pm 0.021$
& $0.045 \pm 0.006$
& $0.033 \pm 0.005$
& $0.067 \pm 0.050$
& $0.040 \pm 0.008$
& $0.829 \pm 0.047$ \\
Low-latent OTD
& 32
& $0.143 \pm 0.008$
& $0.045 \pm 0.002$
& $0.034 \pm 0.001$
& $0.087 \pm 0.066$
& $\mathbf{0.037 \pm 0.000}$
& $0.809 \pm 0.067$ \\
\midrule
Conditional OT
& 1536
& $0.169 \pm 0.007$
& $0.078 \pm 0.005$
& $0.050 \pm 0.002$
& $0.130 \pm 0.031$
& $0.142 \pm 0.003$
& $0.968 \pm 0.014$ \\
Conditional OT
& 512
& $0.156 \pm 0.013$
& $0.059 \pm 0.011$
& $0.041 \pm 0.006$
& $0.076 \pm 0.013$
& $0.089 \pm 0.021$
& $0.937 \pm 0.002$ \\
Conditional OT
& 128
& $0.159 \pm 0.007$
& $0.058 \pm 0.004$
& $0.040 \pm 0.001$
& $0.058 \pm 0.005$
& $0.078 \pm 0.015$
& $0.876 \pm 0.057$ \\
Conditional OT
& 32
& $0.146 \pm 0.013$
& $0.048 \pm 0.005$
& $0.036 \pm 0.005$
& $0.069 \pm 0.083$
& $0.043 \pm 0.003$
& $0.829 \pm 0.081$ \\
\bottomrule
\end{tabular}%
}
\end{table}

The results show two different effects of reducing the transported latent
dimension. For the OTD architecture, matching the original mask dimension remains best in terms of pointwise predictive accuracy: $d_z=1536$ gives the lowest RMSE, MAE, and energy score. However, the reduced latent models remain close despite using substantially smaller transport spaces. In particular, $d_z=512$ obtains RMSE $0.1397 \pm 0.0025$ and $S_{\text{ES}}$ $0.0330 \pm 0.0013$, compared with $0.1341 \pm 0.0026$ and $0.0310 \pm 0.0003$ for $d_z=1536$. The smallest model, $d_z=32$, gives the sharpest intervals, but at the cost of worse calibration error.

The conditional formulation behaves differently. There, reducing $d_z$ improves the predictive metrics relative to the matched $d_z=1536$ model, with $d_z=32$ giving the best RMSE, MAE, and $S_{\text{ES}}$ among the conditional models. This suggests that the concatenated-latent conditional field does not benefit from a large transported code in the same way as the mask-based OTD architecture. The computational effect is also much stronger in the conditional model: the matched conditional model requires approximately $0.952$ seconds per epoch, whereas $d_z=512$, $d_z=128$, and $d_z=32$ require approximately $0.333$, $0.173$, and $0.153$ seconds per epoch, respectively. In contrast, the low-latent OTD runs all take about $0.185$--$0.189$ seconds per epoch, so reducing $d_z$ mainly reduces the transport dimension rather than the full wall-clock cost.

Calibration coverage reflects this trade-off. Overall, the mask-based low-latent OTD model is the stronger predictive architecture in this experiment. The matched $d_z=1536$ version is best overall, but $d_z=512$ and $d_z=128$ provide competitive lower-dimensional alternatives. The conditional model is more sensitive to the choice of $d_z$: large latent codes are costly and over-dispersed, while smaller codes improve pointwise metrics and sharpness but lose coverage.

The relevant structural observation remains that, in the original OTD model, the transport dimension is tied to the predictive architecture. With the new low-latent OTD parameterisation, $d_z$ becomes a free hyperparameter while preserving the original mask-based inductive bias.

\end{document}